\documentclass[11pt]{article}

\usepackage[T1]{fontenc}
\usepackage[utf8]{inputenc}
\usepackage{newtxtext,newtxmath}
\usepackage[scaled=0.92]{helvet}
\usepackage{microtype}
\usepackage[a4paper,top=2.4cm,bottom=2.4cm,left=2.6cm,right=2.6cm]{geometry}
\usepackage{graphicx}
\graphicspath{{./}}
\usepackage{xcolor}
\usepackage{booktabs}
\usepackage{array}
\usepackage{tabularx}
\usepackage{multirow}
\usepackage{makecell}
\usepackage{adjustbox}
\usepackage{caption}
\usepackage{subcaption}
\usepackage{float}
\usepackage[section]{placeins}
\usepackage{enumitem}
\usepackage{titlesec}
\usepackage{amsmath}
\usepackage{wrapfig}
\usepackage[most]{tcolorbox}
\usepackage[colorlinks=true,allcolors=accent]{hyperref}

\definecolor{accent}{HTML}{2B5C8A}
\definecolor{ink}{HTML}{1f2430}
\definecolor{absframe}{HTML}{9bb4cc}
\definecolor{absback}{HTML}{f6f9fc}
\definecolor{good}{HTML}{3c7a3c}
\definecolor{warn}{HTML}{9a5b1e}
\definecolor{rulecol}{HTML}{cfd6dd}

\titleformat{\section}{\sffamily\bfseries\large\color{accent}}{\thesection}{0.6em}{}
\titleformat{\subsection}{\sffamily\bfseries\normalsize\color{accent!85!black}}{\thesubsection}{0.5em}{}
\titleformat{\subsubsection}{\sffamily\bfseries\small\color{ink}}{\thesubsubsection}{0.5em}{}
\titlespacing*{\section}{0pt}{11pt}{4pt}
\titlespacing*{\subsection}{0pt}{8pt}{3pt}

\newtcolorbox{abstractbox}{
  enhanced, breakable, colback=absback, colframe=absframe,
  boxrule=0.8pt, arc=4pt, left=14pt, right=14pt, top=10pt, bottom=10pt,
  before skip=8pt, after skip=10pt
}

\newcommand{\lead}[1]{\smallskip\noindent\textbf{#1}\ }

\newcolumntype{C}[1]{>{\centering\arraybackslash}p{#1}}
\newcolumntype{Y}{>{\centering\arraybackslash}X}

\makeatletter
\newcommand{\bstctlcite}[1]{%
  \@for\@citeb:=#1\do{%
    \if@filesw\immediate\write\@auxout{\string\citation{\@citeb}}\fi}}
\makeatother

\setlist[itemize]{leftmargin=1.4em,itemsep=2pt,topsep=3pt}
\setlist[enumerate]{leftmargin=1.6em,itemsep=2pt,topsep=3pt}
\setlist[description]{leftmargin=1.8em,itemsep=3pt,topsep=3pt}

\begin{document}
\bstctlcite{BSTcontrol}
\pagestyle{plain}

\begin{center}
{\sffamily\bfseries\color{accent}\small ROBOTICS · TECHNICAL REPORT}\\[2pt]
{\color{rulecol}\rule{\linewidth}{1pt}}\\[14pt]
{\sffamily\bfseries\fontsize{19}{24}\selectfont
Assessing the Marginal Utility of Exocentric Head Views in UMI-Style Large-FoV Wrist-Camera Manipulation\par}
\vspace{5pt}
{\sffamily\large A Controlled Study for Tabletop Bimanual Manipulation\par}
\vspace{10pt}
{\normalsize Public Technical Report\quad Date: October 2026\par}
\vspace{3pt}
{\small
Yuhui Duan\textsuperscript{1}\quad
Shi Jin\textsuperscript{1}\quad
Yuntian Wang\textsuperscript{1}\\[2pt]
Jin Wang\textsuperscript{1}\quad
Siao Liu\textsuperscript{1}\quad
Wenchang Gao\textsuperscript{2,*}\quad
Zhaxizhuoma\textsuperscript{3,*}\quad
Yan Ding\textsuperscript{4,5}\par}

\vspace{3pt}

{\footnotesize
\textsuperscript{1}Soochow University\quad
\textsuperscript{2}Independent Researcher\quad
\textsuperscript{3}Shanghai Jiao Tong University\quad
\textsuperscript{4}Fudan University\\[1pt]
\textsuperscript{5}Xi'an Jiaotong-Liverpool University\par}

\vspace{3pt}

{\footnotesize
\textsuperscript{*}Corresponding authors:
\texttt{wenchanggao93@gmail.com},
\texttt{zxzmkufufu@gmail.com}\par}
\vspace{2pt}
%{\small\itshape Real-robot tasks: tabletop bimanual manipulation\quad|\quad View configurations: wrist cameras / head view\par}
%{\small\itshape Policy backbone: OpenPI $\pi_{0.5}$\par}
{\color{rulecol}\rule{\linewidth}{0.6pt}}
\end{center}

\begin{abstractbox}
\begin{center}{\sffamily\bfseries\color{accent}Abstract}\end{center}
\noindent
Multi-view perception is often treated as a safe design choice for robot manipulation: if one camera is useful, more cameras should be better.
We revisit that assumption in tabletop bimanual manipulation by comparing policies trained with two large-field-of-view (FoV) wrist cameras against policies that additionally receive a head view, using controlled-variable experiments on a real bimanual platform.
Across three tabletop tasks, adding a head view does not consistently improve task completion or execution quality: when the wrist cameras already observe the objects, grippers, and goal regions, wrist-only policies form a strong baseline.
Restricting the wrist-camera FoV increases the potential value of a head view, but performance improves most reliably when the added view supplies concentrated, task-relevant information: region-of-interest (ROI) cropping yields more consistent head-view gains than the uncropped image, which is dominated by background and task-irrelevant regions.
The data-scaling results further suggest an interaction between FoV and demonstration count: wide observations improve coverage but may require more data to stabilize, and they may reach a higher performance ceiling once enough demonstrations are available.
These results suggest that camera design should be framed as a task-relevance and information-density problem rather than as a simple count of input views.
\end{abstractbox}

\noindent\textbf{Keywords:}\ tabletop manipulation; bimanual manipulation; wrist camera; head view; field of view; view selection; action sensitivity.

\section{Introduction}

Multi-view visual input is a common choice in robot manipulation.
A third-person or head-mounted camera can provide a stable scene reference, while wrist cameras provide close-up observations of the gripper, object, and contact region.
This intuition motivates a simple design rule: adding a head view should improve policy performance because it gives the model more information.

\begin{figure}[!t]\centering
\includegraphics[width=0.98\linewidth]{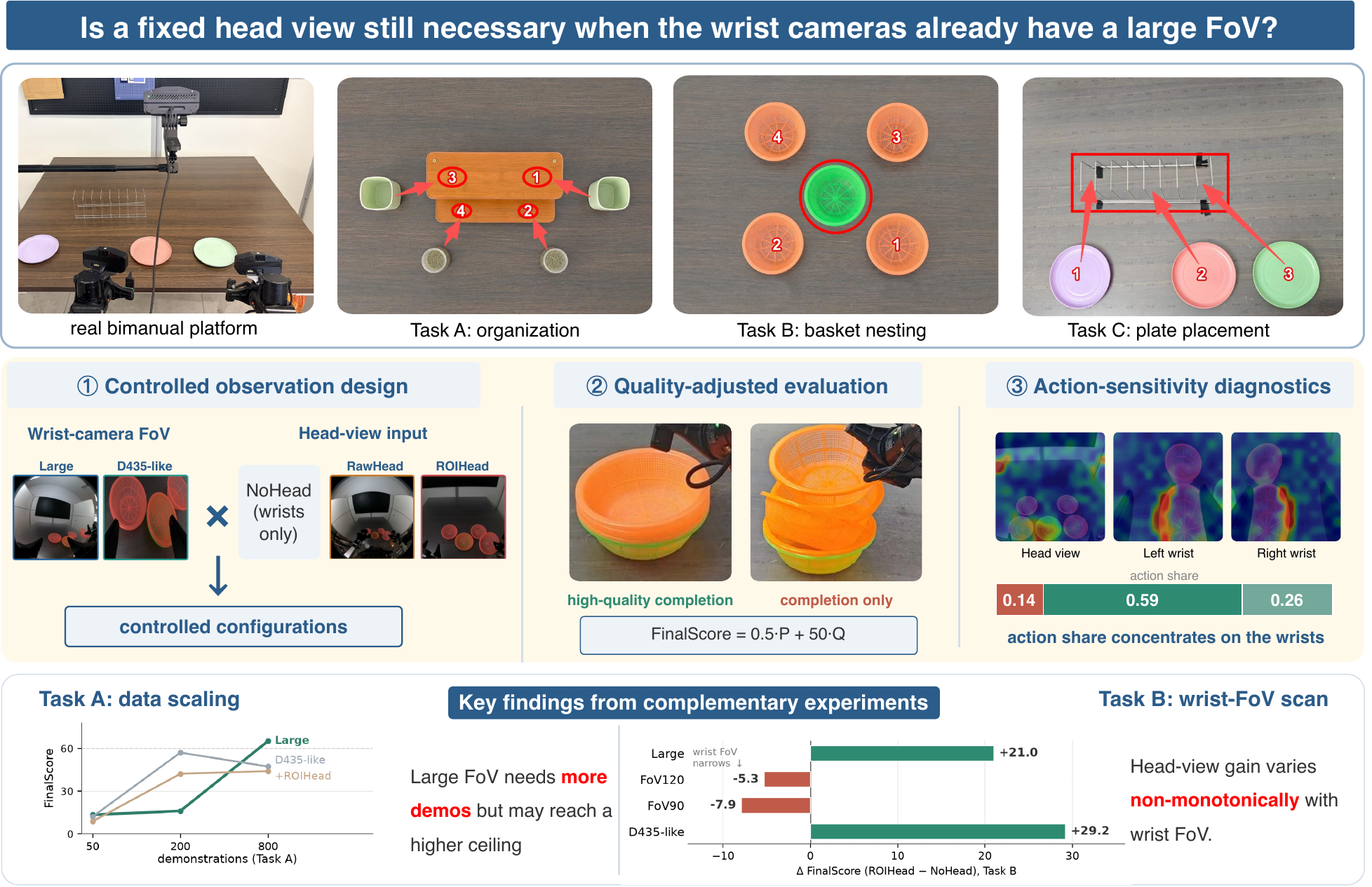}
\caption{Overview of the study, organized by its three contributions and its key findings. Top: the guiding question and the experimental setting---a real bimanual platform with two large-FoV wrist cameras and one fixed head camera, evaluated on three tabletop tasks. \textcircled{\scriptsize 1} The controlled observation design crosses wrist-camera FoV (Large, D435-like; shown by representative wrist frames) with head-view input (NoHead, RawHead, ROIHead), yielding the controlled configurations G1--G6 evaluated in Sec.~\ref{sec:experiments}. \textcircled{\scriptsize 2} Each rollout is scored with a quality-adjusted metric that separates high-quality success from coarse completion, illustrated by two Task B final states. \textcircled{\scriptsize 3} RISE-style action-sensitivity diagnostics: in the Task B example the wrist views carry most of the step-level action share ($0.59+0.26$ versus $0.14$ for the head view), which contributes low-weight global context. Bottom band: key findings from the two complementary experiments---the Task A data-scaling trend, which suggests that large-FoV input needs more demonstrations but may reach a higher performance ceiling, and the Task B wrist-FoV scan, where the ROIHead-versus-NoHead FinalScore contrast is \emph{non-monotonic} in wrist FoV ($+21.0$ Large, $-5.3$ FoV120, $-7.9$ FoV90, $+29.2$ D435-like).}
\label{fig:teaser}
\end{figure}

In practice, this rule is often followed by default: many manipulation systems include a fixed third-person or head camera in their standard observation space~\cite{brohan2022rt1,chi2023diffusion,zhao2023act,kim2024openvla,octo2024}, and the choice is usually inherited as a data-collection convention rather than supported by controlled comparisons or uncertainty analysis.
At the same time, systems built on the Universal Manipulation Interface (UMI) show that wrist-mounted cameras alone can support a wide range of real-world manipulation tasks~\cite{levine2016handeye,chi2024umi}.
This contrast motivates the doubt examined in this study: an extra head view is not automatically informative---it can duplicate what large field-of-view (FoV) wrist cameras already observe, spend fixed image resolution or visual-token capacity on background, and add sensor, calibration, and data-collection cost~\cite{zeng2020transporter,shridhar2021cliport}.

Our controlled experiments show that this default choice is incomplete. 
The head view is most useful when it adds information that the wrist cameras do not already provide, and when that information is concentrated enough for the policy to exploit.
In tabletop bimanual manipulation, however, two large-FoV wrist cameras already provide sufficient coverage of most task-relevant state information: the target object, the end effectors, and the goal region.
In this condition, a head view is not a necessary input channel for current task, and it does not consistently improve success or execution quality.
In some cases, especially when the head image is broad and uncropped, the extra view introduces background clutter, occlusion, or a lower density of task-relevant pixels.

The relevant distinction is not a comparison between a two-camera configuration versus a three-camera configuration; it is a comparison between expanded visual coverage and supplementary usable task evidence. 
This distinction has practical consequences for both demonstration collection and deployment. 
Increasing FoV, or adding an uncropped head stream, seem like a conservative design choice because make it unlikely to miss the table layout. 
However, under a fixed image resolution or visual-token budget, the extra coverage also spends capacity on walls, table boundaries, robot links, and other task-irrelevant contexts. 
This can reduce the density of task-relevant pixels, increase the amount of data needed for the policy to learn what to ignore, and even make the learned policy more sensitive to background or camera-placement changes.
The opposite failure mode is also important. 
A narrower wrist view can concentrate the gripper-object contact region, but it may lose the goal region or global layout as the arm moves. 
In that regime, a head view can become useful, not because it is provides an extra viewpoint, but because it supplies missing layout information that the wrist views do not provide. 
The design question is therefore not whether a head view is universally good or bad, but when does it provide high-density, low-interference incremental evidence that improves execution quality without slowing down learning or reducing robustness.

Figure~\ref{fig:teaser} summarizes the study and previews its three contributions---the controlled two-factor comparison, the quality-adjusted evaluation, and the action-sensitivity diagnostics---together with its key finding: the head-view contrast is not a monotonic improvement in large-FoV wrist FoV settings.
Concretely, we run three controlled experiments on a real bimanual platform with a fixed $\pi_{0.5}$ policy backbone.
The main experiment compares six visual-input configurations (G1--G6), which cross wrist-camera FoV with head-view type, on three tabletop tasks: object organization, basket nesting, and plate placement; every task is evaluated under three fixed initial layouts (Init01--Init03) so that initial-state coverage is identical across configurations.
A dedicated wrist-FoV scan on the basket-nesting task then measures how the head-view contrast changes as wrist FoV is progressively restricted.
Finally, a data-scaling experiment on the object-organization task trains representative configurations with 50, 200, and 800 demonstrations to test how visual-input design interacts with data scale.
Separately, an unscored toy-transfer diagnostic illustrates a boundary condition in which the initial wrist views omit the target location while the head view reveals it.
Across these experiments, large-FoV wrist-only policies form a strong baseline. 
A region-of-interest (ROI) cropped head view helps mainly when wrist visibility is restricted or the task depends heavily on global localization.
A raw head view is associated with lower quality-adjusted performance in several contrasts.
The data-scaling results suggest that large-FoV input needs more demonstrations to stabilize, as its advantage appears only at the largest scale we test.

\paragraph{Research questions.} We study four questions:
\begin{description}
  \item[RQ1] How much additional performance gain does a head view provide when large-FoV wrist cameras already observe the task-critical regions?
  \item[RQ2] Does restricted wrist FoV increase the potential value of a head view, and when does that potential fail to become a higher quality-adjusted score?
  \item[RQ3] Does a cropped head view focusing on the task region provide denser, lower-interference information than a raw wide-angle head view?
  \item[RQ4] How does the number of available training demonstrations influence the effect of visual-input configuration?
\end{description}

\paragraph{Contributions.} This study makes three practical contributions.
First, it provides a controlled real-robot comparison that separates two visual-input factors: wrist-camera FoV and head-view type. We compare wrist-only baselines with wrist-plus-RawHead and wrist-plus-ROIHead policies across multiple wrist-FoV settings.
Despite being useful when wrist visibility is restricted, the head camera is often redundant once the wrist cameras already cover the task-critical regions: under the Large-FoV setting, adding a raw head view lowers the mean FinalScore across the three tasks (64.7 vs.\ 70.4 for the wrist-only baseline), and the head-view contrast on Task B is non-monotonic as wrist FoV shrinks ($+21.0$ under Large, $-5.3$ and $-7.9$ at the medium settings, $+29.2$ under D435-like).
Second, beyond binary task success, it evaluates each rollout with a quality-oriented metric suite, including completion rate, quality-adjusted FinalScore, Quality Success Rate, localization rate, retry penalty, and step count. This allows us to distinguish mere task completion from accurate, stable, and efficient execution.
Third, we combine rollout results with action-sensitivity diagnostics to analyze view roles, suggesting that dense task-region evidence matters more than simply adding cameras.

\section{Related Work}

\subsection{Demonstration Collection and Generalist Robot Policies}
Recent robot learning has moved from single-task policies toward scalable vision-language-action models and large-scale imitation learning: RT-1, RT-2, and PaLM-E show that large-scale robot data and visual-language knowledge can support real-world control~\cite{brohan2022rt1,brohan2023rt2,driess2023palme}; Open X-Embodiment, OpenVLA, and Octo advance cross-embodiment data scale and open-source reproducibility~\cite{openx2023,kim2024openvla,octo2024}; $\pi_0$ and $\pi_{0.5}$ emphasize large-scale pretraining followed by downstream adaptation~\cite{black2024pi0,pi052025}; and BC-Z and R3M show that data diversity and visual representations shape downstream generalization~\cite{jang2022bcz,nair2022r3m}.

To isolate camera-view effects, we fix $\pi_{0.5}$ from the OpenPI series as the policy backbone.
We then study how visual input design affects tabletop manipulation under the same training and inference pipeline, rather than comparing the upper limits of different backbones.
This setting shifts the question from whether the model is strong enough to which camera views provide task-relevant information.

\subsection{UMI, Wrist Views, and Eye-in-Hand Manipulation}
Wrist cameras and eye-in-hand observations are widely used in manipulation because they provide local visual signals near the gripper, object, and contact region, a spatial relation that early large-scale hand-eye coordination work already identified as central to closed-loop grasping~\cite{levine2016handeye}.
Controlled studies further show that hand-centric observations can improve policy training and generalization relative to purely third-person viewpoints~\cite{hsu2022seeing}.
Diffusion Policy and ACT improve real-robot imitation learning through action diffusion modeling and action chunking~\cite{chi2023diffusion,zhao2023act}. 
Robot-agnostic policies and the Universal Manipulation Interface (UMI) converts portable EEF actions into deployable robot execution through inverse kinematics~\cite{fang2025flex,gao2026step,chi2024umi}. 
UMI-Bench standardizes real-robot evaluation for UMI-style wrist-view policies by aligning demonstration collection, scene reset, and policy execution in one reproducible protocol with partial-credit scoring~\cite{jin2026umibench}.

These systems directly motivate the setting of this study.
If UMI-style large-FoV wrist cameras already cover most task-critical information, the value of an additional head view should not be assumed.
It should be tested under controlled wrist-FoV, head-view, and data-scale conditions.
We therefore treat large-FoV dual wrist cameras as a strong baseline, not as an incomplete setup that always requires a head view.

\subsection{Multi-View Perception, Head Views, and View Redundancy}
Multi-view observations are often used to reduce occlusion, provide complementary spatial information, and improve global scene understanding; in robot learning, a third-person or exocentric view offers a stable scene reference when hand-centric views cannot see the target or goal region.
However, adding a camera does not automatically add useful information: an extra view dominated by background regions, static table areas, or task-irrelevant objects can dilute task-relevant signals and make feature learning harder.
Transporter Networks, CLIPort, and PerAct show that spatial structure, language conditioning, and 3D or multi-view representations can improve data efficiency and generalization in tabletop manipulation~\cite{zeng2020transporter,shridhar2021cliport,shridhar2022peract}; multi-view world-model pretraining likewise exploits complementary viewpoints when several cameras are available~\cite{seo2023multiview}; and active perception work frames viewpoint selection as a design variable coupled with task-state estimation and action decisions~\cite{zaky2020activeperception,fang2025flex}.

This study therefore asks a narrower question.
It asks whether a fixed head view is still necessary for tabletop manipulation when large-FoV wrist cameras already cover the interaction region.
We separate wrist FoV from head-view design and compare NoHead, RawHead, and ROIHead under Large, FoV120, FoV90, and D435-like wrist-FoV settings.
This design tests whether head-view contrasts remain consistent under the reported descriptive uncertainty as wrist information becomes restricted, and whether any positive effect is task-dependent.

\subsection{ROI Cropping and Visual Attribution}
The presence of a view does not mean that the policy uses it effectively.
A raw head view may include the target region, but it may also include background, robot body parts, table boundaries, and other irrelevant content.
ROI cropping and view selection increase the density of task-relevant information and can reduce interference from irrelevant pixels.

In this study, we additionally adapt randomized input masking in the style of RISE (Randomized Input Sampling for Explanation)~\cite{petsiuk2018rise} as an action-sensitivity diagnostic; its formulation is given in Section~\ref{sec:action-sensitivity}.

\section{Experimental Setup}
This section describes the tasks, visual-input configurations, experiment design, and evaluation protocol.
Here, a visual-input configuration denotes a controlled observation condition specified by a wrist-camera FoV setting and a head-view type, with the concrete settings defined in Section~\ref{sec:view-configurations}.
We keep the policy backbone and train/inference protocol fixed while varying these visual-input factors and the training data scale, with the goal of systematically analyzing the practical role of head views in tabletop manipulation under controlled observation conditions.
Each task is evaluated under three fixed initial layouts (Init01--Init03) that relocate the task objects between rollouts, so that no configuration can succeed by replaying a single memorized trajectory; the layouts and their design intent are detailed in Section~\ref{sec:tasks}.

\subsection{Tasks}
\label{sec:tasks}
We design three main experimental tasks to cover different manipulation abilities.
They correspond to long-horizon manipulation, bimanual coordination, and fine-grained localization.

\begin{figure}[!ht]\centering
\includegraphics[width=0.92\linewidth]{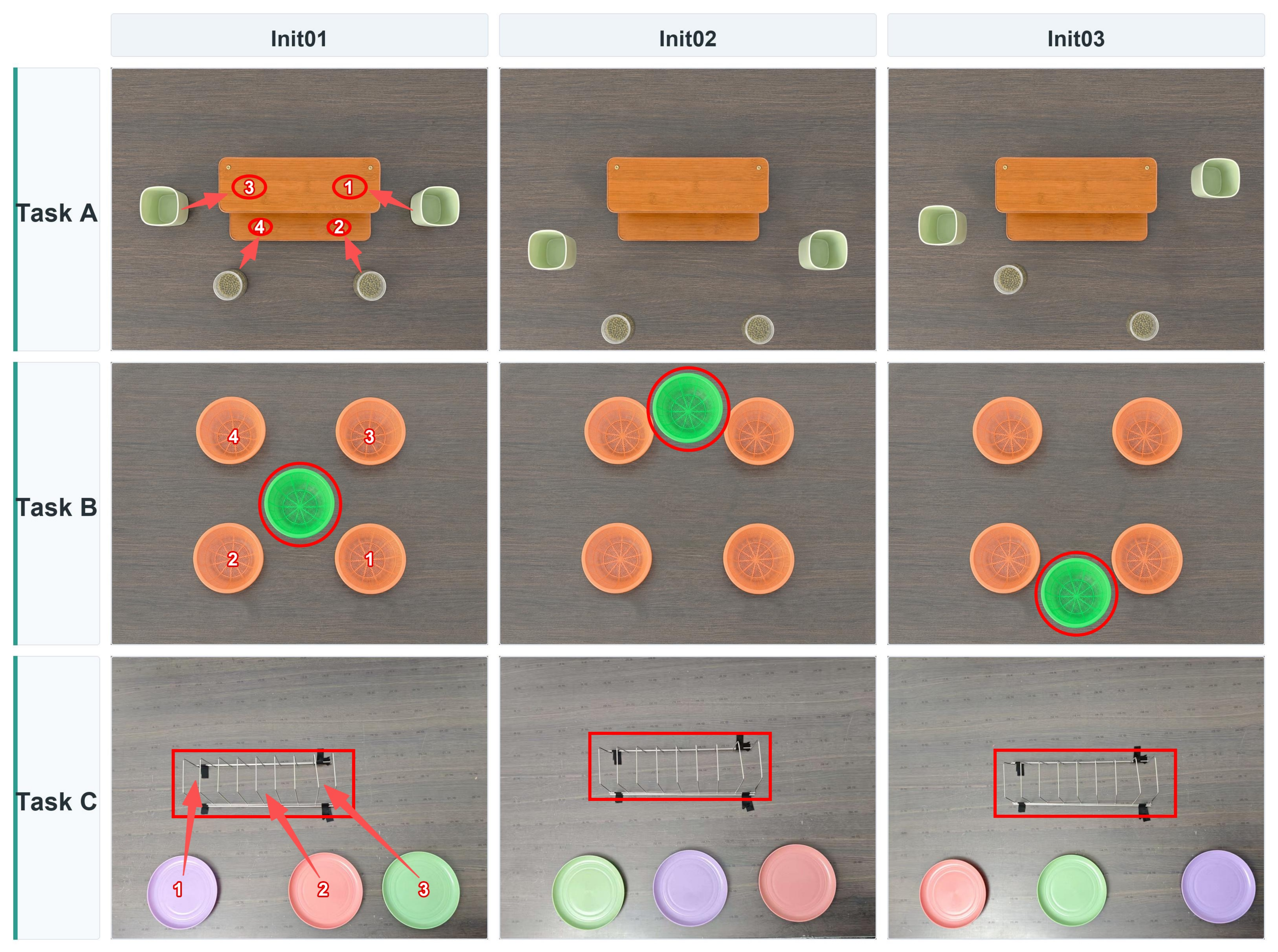}
\caption{Task overview and initial layouts. Rows show Tasks A--C, and columns show Init01--Init03. Red annotations specify the task goals, with numbered assignments illustrated in Init01. In Task A, arrows map the four objects to storage positions 1--4. In Task B, the green target basket is circled, and the surrounding baskets to be nested are labeled 1--4. In Task C, arrows map plates 1--3 to rack slots, and boxes mark the rack region.}
\label{fig:task-overview}
\end{figure}

For each main task, we define three fixed initial layouts, Init01--Init03, illustrated in Fig.~\ref{fig:task-overview}; the red boxes and arrows in the figure mark the task-relevant regions and object-to-goal assignments used in the task descriptions below.
Across Init01--Init03, the object placements in Task A and Task B are changed so that success cannot come from replaying one memorized trajectory, and Task C additionally permutes the relative positions of the colored plates to probe color generalization and instruction following.
All experimental conditions use the fixed Init01--Init03 evaluation schedule to ensure identical initial-state coverage across the experiments.

\lead{Task A: Four-Object Organization}
The robot must sequentially grasp four target objects and place them into a specified storage region; the red arrows in Fig.~\ref{fig:task-overview} map each object to its assigned storage position.
This task includes repeated grasping, transport, and placement.
It is a typical long-horizon manipulation task.
We use this task to evaluate task progress, execution stability, and overall completion efficiency of each configuration.
Since it includes multiple targets and a large spatial range, Task A is also used for the data-scaling experiment.

\lead{Task B: Basket Nesting}
The robot must use both arms to sequentially grasp the four surrounding baskets (labeled 1--4 in Fig.~\ref{fig:task-overview}) and nest them into the central target basket (circled in the figure).
A key challenge is accurate bimanual coordination and relative pose estimation, because each successful nesting depends on the final relative pose between the moved basket and the central target basket.
We therefore use Task B as the main task for the wrist-FoV comparison.
It tests whether the head view provides extra localization information when wrist vision is restricted.

\lead{Task C: Plate Placement}
The robot must grasp target plates and place them accurately into a plate rack; in Fig.~\ref{fig:task-overview}, the red box marks the rack region and the red arrows map each plate to its assigned rack slot.
This task focuses on target localization, grasp stability, and final placement precision.
It is sensitive to local visual information, target re-localization, and contact-region judgment.
It is therefore suitable for evaluating the head view in fine manipulation.

In addition to these three main tasks, we later use a supplementary toy-transfer diagnostic to examine the head view under a specific boundary condition.
It is not part of the G1--G6 main experiment matrix and is therefore described together with its analysis in Section~\ref{sec:supp-diagnostic}.

To avoid conclusions being dominated by a single initial state, all three main tasks are evaluated under fixed layout distributions.
Figure~\ref{fig:task-overview} shows Init01--Init03 for each task.

\subsection{Policy Model and Training Details}
All experiments use the OpenPI $\pi_{0.5}$ policy model as the shared policy backbone.
During the experiments, we keep the model architecture, training pipeline, and inference mechanism fixed.
Only the visual input configuration, wrist FoV, and training data scale are changed.
Thus, the results mainly reflect the effect of visual input design rather than differences in model architecture.

All main experiment configurations use full-parameter supervised fine-tuning (SFT) of the OpenPI $\pi_{0.5}$ backbone for 60,000 optimization steps on 8 A100 GPUs, with global batch size 32.
The optimizer and learning-rate settings are fixed across all configurations and listed in Appendix~\ref{app:training}.
During inference, each policy query produces one action chunk, and the robot executes its 50 control steps at a control frequency of 20\,Hz before the next query.
This training and inference setup is kept the same across all tasks and view configurations.

The comparison is not a simple test of whether an additional camera is beneficial.
We treat visual input configurations as an observation-allocation problem.
The key questions are whether a view captures the visual cues needed for the task, whether the model manages to extract this information, and whether an extra view introduces more semantic information or more task-irrelevant noise.
For this reason, we keep the policy backbone and training protocol fixed, and only change the camera input form, wrist FoV, or data scale.

\subsection{Evaluation Metrics}

To ensure the evaluation reproducible, we separate raw rollout observations from derived metrics.
In the scoring record, only directly observed quantities are manually filled; Table~\ref{tab:metric-definitions} defines these quantities and the symbols used below.
The overall evaluation score, denoted as \textit{FinalScore}, task completion, and high-quality success are then computed from fixed formulas.
\begin{table}[!ht]\centering\small
\caption{Directly observed scoring quantities. Only these quantities are manually recorded for each rollout; all derived scores in this section are computed from them by fixed formulas. Localization-related counts apply to Task B and Task C as indicated.}
\label{tab:metric-definitions}
\begin{tabularx}{\linewidth}{c p{0.30\linewidth} X}
\toprule
\textbf{Symbol} & \textbf{Quantity} & \textbf{Definition}\\
\midrule
$n_g$ & Successful grasps & Times a target object is securely lifted by the gripper.\\
$n_p$ & Correct placements & Target objects placed or inserted into the assigned goal region.\\
$n_s$ & Stable placements & Placements where the object retains its intended pose at the target after release.\\
$n_r$ & Retries & Repeated grasp or alignment attempts after a failed attempt.\\
$T$ & Executed steps & Executed low-level control steps in the rollout.\\
$n_{g,\mathrm{loc}}$, $n_{p,\mathrm{loc}}$ & Localized grasps and placements (Task B) & Grasps and placements executed on the correctly localized baskets.\\
$n_{\mathrm{loc}}$ & First-localization decisions (Task C) & Plates for which the first approach targets the correct plate and rack slot.\\
\bottomrule
\end{tabularx}
\end{table}
The primary performance metric in this report is a quality-adjusted task score.
It separates task progress from strict quality success:
\begin{equation}
\mathrm{FinalScore}_{\tau} = \frac{1}{2}P_{\tau} + 50Q_{\tau},
\label{eq:quality-adjusted-score}
\end{equation}
where $P_{\tau}\in[0,100]$ is a task-normalized outcome score and $Q_{\tau}\in\{0,1\}$ is the high-quality success indicator.
We report the fraction of rollouts with $Q_{\tau}=1$ as the Quality Success Rate ($q_{\mathrm{hq}}$).
This definition prevents a long, unstable, or retry-heavy rollout from being treated as equivalent to a clean completion.
It also keeps localization diagnostics separate from the main task-outcome score, except where localization is part of the high-quality criterion.

Object organization involves four target objects and basket nesting involves four baskets nested into the central target basket, so their outcome scores share the same normalization, while plate placement involves three plates:
\begin{align}
P_{\mathrm{org}} &= 100\left(0.25\frac{n_g}{4}+0.45\frac{n_p}{4}+0.30\frac{n_s}{4}\right),\\
P_{\mathrm{basket}} &= 100\left(0.25\frac{n_g}{4}+0.45\frac{n_p}{4}+0.30\frac{n_s}{4}\right),\\
P_{\mathrm{plate}} &= 100\left(0.35\frac{n_g}{3}+0.65\frac{n_p}{3}\right).
\end{align}
Each score reaches 100 when all task objects are properly grasped, correctly placed, and remain stable after the gripper releases. 
The weights emphasize correct placement while preserving partial credit for grasping and stable completion.
This partial-credit design is consistent with recent real-robot benchmark protocols for UMI-style policies~\cite{jin2026umibench}.
Localization metrics are not awarded credits directly. 
For Task B and Task C, they are reported separately and are used only to determine whether a completed rollout qualifies as high-quality.

The task-specific completion and high-quality criteria used together with Eq.~\ref{eq:quality-adjusted-score} are listed in Table~\ref{tab:scoring-protocol} in Appendix~\ref{app:scoring}.
In brief, completion requires the task objects to reach stable goal states, while the high-quality gate additionally requires a bounded step count together with task-specific conditions---no key failure for Task A, and passed localization with limited retries for Task B and Task C; a key failure denotes an execution error that requires an evident recovery or repeated attempt during the rollout.

The reported Steps denote low-level control steps. 
With an action chunk of size 50, a 1000-step rollout corresponds to 20 policy queries.
Each task-specific step threshold is set to the mean step count of a first-attempt success without any retrying on that task. 
Therefore the high-quality gate rewards rollouts that finish in one pass rather than rollouts that recover through repeated attempts.
Each task also has a maximum step budget ($1500$ for Task A and Task C, $1000$ for Task B). 
A rollout exceeding its corresponding budget is cut off at that step before completion. 
Therefore the efficiency of the model cannot be inferred from Mean Steps alone, completion rate should be taken into account as well.

\subsection{Reproducibility Protocol}
\label{sec:repro-protocol}
We fix the sources of variation other than the controlled visual factors to ensure comparison fairness across configurations to support reproduction of our results.
Our control on data scale, initial states, seeds, and inference are summarized in Table~\ref{tab:reproducibility-protocol}.
For each task in the main experiment, G1--G6 use the same data scale and the same evaluation initial-state distribution.
The only differences are wrist FoV and head-view input type, which are intended for comparisons in visual factors in Section~\ref{sec:view-configurations}.
For the data-scaling experiment, the training subsets are nested to reduce confounding from different data splits.

\begin{table}[!ht]\centering\scriptsize
\caption{Reproducibility protocol.}
\label{tab:reproducibility-protocol}
\begin{adjustbox}{width=\linewidth}
\begin{tabular}{p{0.15\linewidth} p{0.13\linewidth} p{0.11\linewidth} p{0.22\linewidth} p{0.14\linewidth} p{0.23\linewidth}}
\toprule
\textbf{Experiment Block} & \textbf{Training Data} & \textbf{Same Data?} & \textbf{Seed Protocol} & \textbf{Init Distribution} & \textbf{Evaluation and Inference Details}\\
\midrule
G1--G6 main experiment & Full / 800 demos per task & Yes, within each task & Same training recipe; training seed is not used as the paired comparison variable & Fixed 4/3/3 initial-layout schedule & Same evaluation schedule per task--configuration pair. All models use action chunk $=50$.\\
\midrule
Task B FoV scan & Full / 800 demos & Yes, within Task B & Same training recipe; comparisons are controlled by FoV/head type and init distribution & Same 4/3/3 split & Large, FoV120, FoV90, and D435-like are compared under NoHead and ROIHead.\\
\midrule
Task A data scaling & 50 / 200 / 800 demos & Nested subsets & Same training recipe; data size is the controlled variable & Same 4/3/3 split & Used to test how visual configuration interacts with demonstration count.\\
\bottomrule
\end{tabular}
\end{adjustbox}
\end{table}

\subsection{View Configurations}
\label{sec:view-configurations}
We separate visual input design into two factors: \textbf{wrist-camera FoV} and \textbf{head-view type}.
The wrist cameras mainly provide local visual information around the grippers, target objects, and contact regions.
Wrist-camera FoV is the first controlled visual factor in this study. 
We evaluate four settings—Large, FoV120, FoV90, and D435-like—which progressively restrict the visible workspace from the original UMI-style large-FoV view to a narrow D435-like crop. Their definitions are summarized in Table~\ref{tab:view-factors}(a) in Appendix~\ref{app:view-config}.

The second controlled visual factor is the head-view type, which includes three settings: NoHead, RawHead, and ROIHead. 
Although RawHead and ROIHead are derived from the same physical camera, they contain substantially different proportions of task-relevant image content.
Table~\ref{tab:view-factors}(b) in Appendix~\ref{app:view-config} summarizes these settings.

\begin{figure}[!ht]\centering
\includegraphics[width=0.84\linewidth]{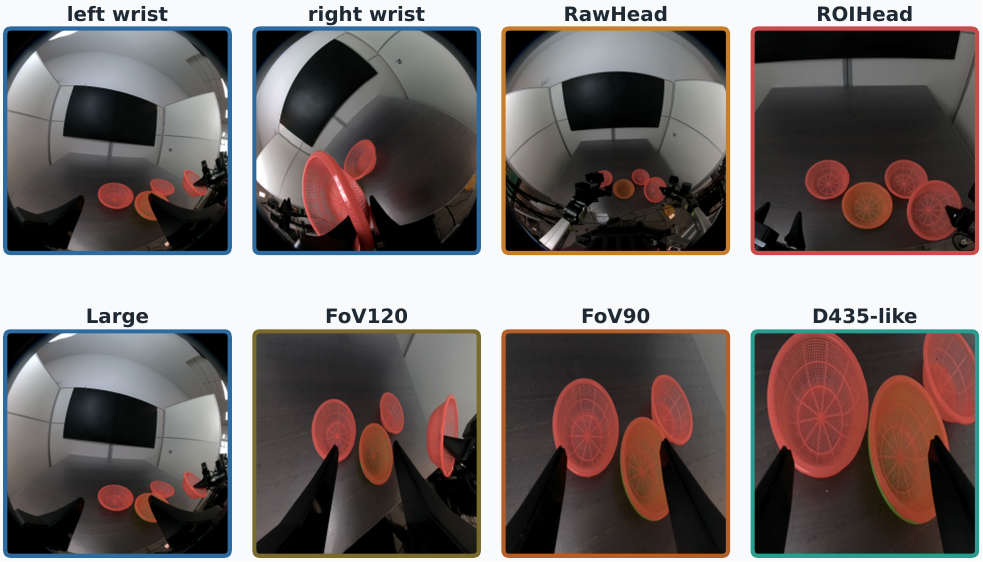}
\caption{Camera configuration schematic using representative policy-input frames from the basket-nesting experiments. Top row: left-wrist, right-wrist, RawHead, and ROIHead inputs. Bottom row: representative left-wrist inputs under Large, FoV120, FoV90, and D435-like settings. The panels come from separate rollouts and therefore illustrate view geometry and task-content density rather than a frame-matched crop comparison.}
\label{fig:camera-config}
\end{figure}

What affects the policy is the relative proportion of targets, grippers, and background in the input images.
Figure~\ref{fig:camera-config} compares representative policy inputs from the basket-nesting experiments: the top row shows the left-wrist, right-wrist, RawHead, and ROIHead views, while the bottom row shows representative left-wrist inputs under the four FoV settings. 
The panels are drawn from the corresponding configurations rather than from a frame-matched recording, so the figure is intended to illustrate view geometry and task-content density rather than a pixel-wise controlled crop comparison.

To characterize the visual information available from each input view, we inspect representative observation frames and record the visibility of the target object, gripper, and goal region. 
We also qualitatively assess the proportion of task-relevant pixels and background content using three relative levels: Low, Medium, and High. 
This audit is used to support the interpretation of the view-design results rather than serving as an additional performance metric. 
The resulting observations are summarized in Table~\ref{tab:view-visibility} in Appendix~\ref{app:view-config}.
It gives a clear context for why different views may provide different localization and execution-quality cues.

This audit illustrates the main visibility interpretation used in this study.
The number of cameras is not the determining factor.
Large Wrist has no head view, but the target, gripper, and contact region are more concentrated in the wrist images, so it forms a strong baseline.
RawHead sees the object and goal region, but the target occupies a only small image area, and the background ratio is high.
This can lead to low-quality completion or longer trajectories to completion.
ROIHead can be easier for the policy to use because task-relevant regions are more concentrated.
The main risk of D435-like Wrist is that the target or goal region can be cropped out at critical moments.

Based on these two factors, we construct six main experiment configurations, denoted as G1--G6 in Table~\ref{tab:g-configs}.
The D435-like setting only simulates a smaller FoV by image cropping.
It is not equivalent to the full imaging characteristics of a physical Intel RealSense D435 camera.

\begin{table}[!ht]\centering\scriptsize
\caption{Six main experiment configurations.}
\label{tab:g-configs}
\begin{tabularx}{\linewidth}{c c c c Y}
\toprule
\textbf{ID} & \textbf{Wrist FoV} & \textbf{Head Type} & \textbf{Input Configuration} & \textbf{Purpose}\\
\midrule
G1 & Large & NoHead & Large Wrist Only & Large-FoV dual wrist baseline.\\
G2 & Large & RawHead & Large Wrist + RawHead & Analyze RawHead under the large-FoV wrist setting.\\
G3 & Large & ROIHead & Large Wrist + ROIHead & Analyze ROIHead under the large-FoV wrist setting.\\
G4 & D435-like & NoHead & Cropped Wrist Only & Simulate restricted wrist vision.\\
G5 & D435-like & RawHead & Cropped Wrist + RawHead & Analyze RawHead under the small-FoV wrist setting.\\
G6 & D435-like & ROIHead & Cropped Wrist + ROIHead & Analyze ROIHead under the small-FoV wrist setting.\\
\bottomrule
\end{tabularx}
\end{table}

\subsection{Action-Sensitivity Diagnostic}
\label{sec:action-sensitivity}
To probe whether the predicted action is sensitive to perturbing a given view, we use RISE-style randomized input masking as a diagnostic tool~\cite{petsiuk2018rise}.
Random binary masks occlude local regions of a single view while the rest of the observation is unchanged; the induced normalized deviation of the predicted action chunk defines a cell-level action-sensitivity score $S_v(c)$, and a view-level share $q_v$ summarizes how much each camera contributes at the selected step.
The full formulation is given in Appendix~\ref{app:rise}.
The displayed heatmaps are normalized only for visualization; optional smoothing, clipping, or alpha scaling does not change $S_v(c)$ or $q_v$.
Note that these maps are not Transformer attention maps thus should not be interpreted as rollout-success evidence.
In this study, they provide diagnostic clues about how different views may contribute to policy decisions.

\section{Experiments and Results}
\label{sec:experiments}
This section presents each experiment together with its results: the G1--G6 main experiment on Tasks A--C, the wrist-FoV scan on Task B, and the data-scaling study on Task A, followed by a supplementary diagnostic case.
Cross-cutting analyses of the mechanisms behind these results---the strength of the large-FoV wrist baseline, the role of head-view input form, view-level attribution, and typical failure modes---are presented in Section~\ref{sec:analysis}.

\subsection{Main Experiment}

The main experiment evaluates the six visual configurations G1--G6 on Task A, Task B, and Task C.
Each task-configuration pair uses the same fixed evaluation schedule.
All quantitative results are recomputed from per-rollout scoring records.
When uncertainty summaries are shown, we bootstrap over the matched rollout slots within each comparison while preserving the fixed initial-layout schedule.
These intervals should be read as descriptive uncertainty rather than formal significance tests.

Rather than walking through all pairwise comparisons, the analysis in Section~\ref{sec:analysis} reads these results through a small set of contrasts, each mapped to the research question it serves (Table~\ref{tab:main-comparisons}).

We first report the overall results on the three main tasks.
Unless stated otherwise, all main experiment results use the same fixed evaluation schedule described above.

Overall, large-FoV wrist cameras already form a strong baseline, and adding a head view does not produce a consistent performance improvement across all tasks under the reported descriptive uncertainty.

The overall conclusion is based on the full set of metrics across the three main tasks, not only on FinalScore.
Table~\ref{tab:main-results-overview} reports completion rate $p_{\mathrm{comp}}$, FinalScore, its score split, Quality Success Rate $q_{\mathrm{hq}}$, and average Steps for G1--G6.

\begin{table}[!ht]\centering\scriptsize
\caption{Main G1--G6 results for Task A/B/C. The table reports completion rate $p_{\mathrm{comp}}$, FinalScore, Quality Success Rate $q_{\mathrm{hq}}$, and Steps. The Score Split column reports outcome contribution plus high-quality bonus, which sum to FinalScore. Reporting both $p_{\mathrm{comp}}$ and $q_{\mathrm{hq}}$ separates task non-completion from completed but low-quality rollouts.}
\label{tab:main-results-overview}
\begin{adjustbox}{width=\linewidth}
\begin{tabular}{c c c c c c c c c}
\toprule
Task & Config & Wrist FoV & Head Type & \makecell{$p_{\mathrm{comp}}$\\Completion} & FinalScore & \makecell{Score\\Split} & \makecell{$q_{\mathrm{hq}}$\\Quality\\Success Rate} & Steps\\
\midrule
Task A & G1 & Large & NoHead  & 50\%  & 65.2 & $45.2+20.0$ & 40\% & 975.0\\
Task A & G2 & Large & RawHead & 90\%  & 49.2 & $49.2+0.0$ & 0\%  & 1045.0\\
Task A & G3 & Large & ROIHead & 80\%  & 59.2 & $49.2+10.0$ & 20\% & 1050.0\\
Task A & G4 & D435-like & NoHead  & 60\% & 47.3 & $42.3+5.0$ & 10\% & 1200.0\\
Task A & G5 & D435-like & RawHead & 70\% & 45.1 & $45.1+0.0$ & 0\%  & 1200.0\\
Task A & G6 & D435-like & ROIHead & 30\% & 44.1 & $44.1+0.0$ & 0\%  & 1165.0\\
\midrule
Task B & G1 & Large & NoHead  & 60\% & 68.2 & $43.2+25.0$ & 50\% & 837.5\\
Task B & G2 & Large & RawHead & 50\% & 67.2 & $47.2+20.0$ & 40\% & 905.0\\
Task B & G3 & Large & ROIHead & 80\% & 89.2 & $49.2+40.0$ & 80\% & 790.0\\
Task B & G4 & D435-like & NoHead  & 20\% & 42.2 & $37.2+5.0$ & 10\% & 945.0\\
Task B & G5 & D435-like & RawHead & 30\% & 57.5 & $42.5+15.0$ & 30\% & 947.5\\
Task B & G6 & D435-like & ROIHead & 50\% & 71.4 & $46.4+25.0$ & 50\% & 907.5\\
\midrule
Task C & G1 & Large & NoHead  & 80\%  & 77.8 & $47.8+30.0$ & 60\% & 1100.0\\
Task C & G2 & Large & RawHead & 80\%  & 77.8 & $47.8+30.0$ & 60\% & 1095.0\\
Task C & G3 & Large & ROIHead & 100\% & 90.0 & $50.0+40.0$ & 80\% & 1040.0\\
Task C & G4 & D435-like & NoHead  & 0\% & 28.0 & $28.0+0.0$ & 0\%  & 1365.0\\
Task C & G5 & D435-like & RawHead & 0\% & 36.9 & $36.9+0.0$ & 0\%  & 1260.0\\
Task C & G6 & D435-like & ROIHead & 90\% & 58.3 & $48.3+10.0$ & 20\% & 1240.0\\
\bottomrule
\end{tabular}
\end{adjustbox}
\end{table}

\begin{figure}[!ht]\centering
\includegraphics[width=0.96\linewidth]{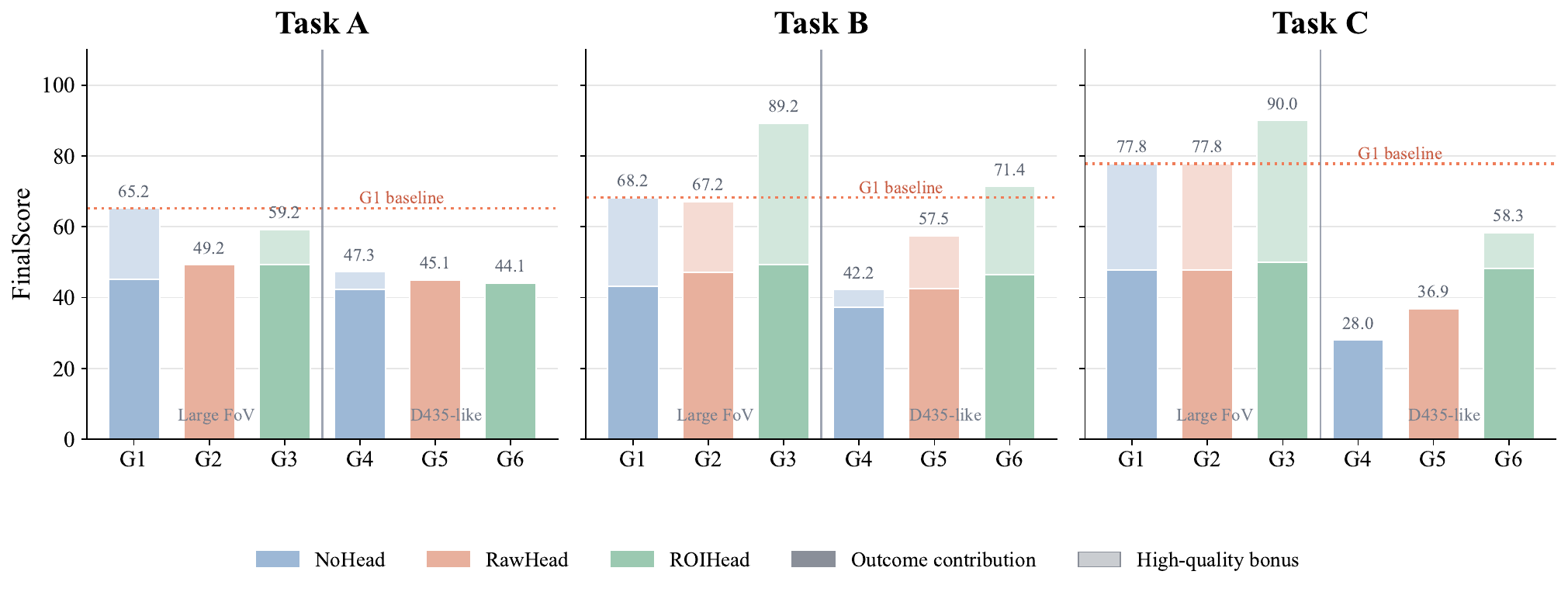}
\caption{G1--G6 bar plots for the three tasks. Each bar shows FinalScore split into the outcome contribution (solid) and the high-quality bonus (light top segment), following the Score Split column in Table~\ref{tab:main-results-overview}. Bars without a light segment complete rollouts without earning any high-quality bonus.}
\label{fig:main-results-bars}
\end{figure}

\begin{figure}[!ht]\centering
\includegraphics[width=0.82\linewidth]{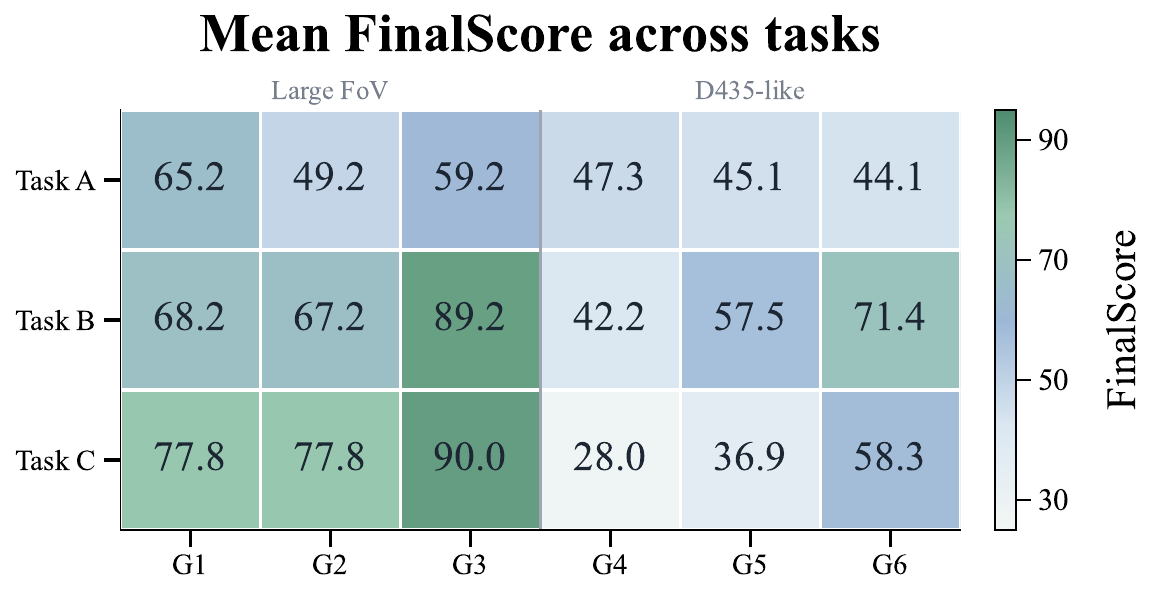}
\caption{Summary matrix of mean FinalScore across tasks and configurations. The continuous heatmap uses a muted blue-green palette consistent with the style of Figure~\ref{fig:main-results-bars}; lighter cells indicate lower FinalScore and darker cells indicate higher FinalScore.}
\label{fig:main-results-heatmap}
\end{figure}

All later contrast analyses are based on the values in Table~\ref{tab:main-results-overview}.
The cross-task pattern is difficult to see from the table alone.
We therefore plot FinalScore by task in Figure~\ref{fig:main-results-bars}.
We also summarize the mean FinalScore pattern across tasks and configurations in Figure~\ref{fig:main-results-heatmap}.

\paragraph{Overall reading.}

The overall results show that the head view is not a stable or universal source of model-performance improvement.
Its measured effect depends on the task type, wrist FoV, and head-view input form.
When the wrist cameras already cover the target object, grippers, and goal region, wrist-only input already forms a strong baseline, and the head view is not a default requirement.
When wrist FoV is restricted, the head view has greater opportunity to provide missing layout or localization evidence, but this opportunity only improves FinalScore when the added information is cleanly converted into high-quality execution.
Positive mean differences should therefore be interpreted together with quality, step count, and uncertainty.
In some tasks, the head view changes localization quality, execution efficiency, or Quality Success Rate.
It does not necessarily change whether the task can be completed.

Task A is a long-horizon, multi-step task.
The large-FoV dual wrist setup already provides relatively complete task information, so the marginal effect of the head view is limited and not stable across metrics.
Task B has higher requirements for bimanual coordination and relative localization, where ROIHead shows the clearest positive mean contrast.
Task C is more sensitive to first localization and final placement accuracy.
When wrist FoV is restricted, the head view is associated with higher mean performance in some contrasts, especially when the added view is ROI-cropped and task-region focused, but this remains condition-specific rather than a global result.

These results show that the role of the head view cannot be reduced to the idea that one more camera is always better.
More precisely, the head view does not materially affect model performance in a consistent way unless it supplies task-relevant information that is missing from the wrist views and the policy uses that information without degrading execution quality.

\subsection{Wrist-FoV Experiment}

To further analyze how wrist FoV affects the apparent effect of the head view, we conduct a dedicated FoV comparison on Task B.
Task B requires strong bimanual coordination and relative localization.
It is therefore suitable for observing whether the head-view effect changes after wrist vision is restricted.

The experiment compares four wrist-FoV settings: Large, FoV120, FoV90, and D435-like.
Under each FoV condition, we evaluate NoHead and ROIHead.
The measured head-view contrast is defined as
\begin{equation}
\Delta_{\mathrm{head}} =
\mathrm{FinalScore}_{\mathrm{ROIHead}} -
\mathrm{FinalScore}_{\mathrm{NoHead}}.
\label{eq:head-gain}
\end{equation}
If the head view compensates for task information that is missing from the wrist observations, restricted wrist FoV should increase the opportunity for a positive head-view contrast.
This opportunity need not produce a monotonic FinalScore gain: the added view must also integrate cleanly with wrist-centered control and improve completion quality, step count, and retry behavior.
Because FinalScore contains both an outcome term and a strict high-quality bonus, the FoV scan must be decomposed by localization, completion, Quality Success Rate, and step-count metrics rather than read only as a single $\Delta_{\mathrm{head}}$ curve.
We therefore treat the FoV scan as a test of conditional utility, not as proof that a head camera has a guaranteed positive effect.

\begin{figure}[!ht]\centering
\includegraphics[width=0.98\linewidth]{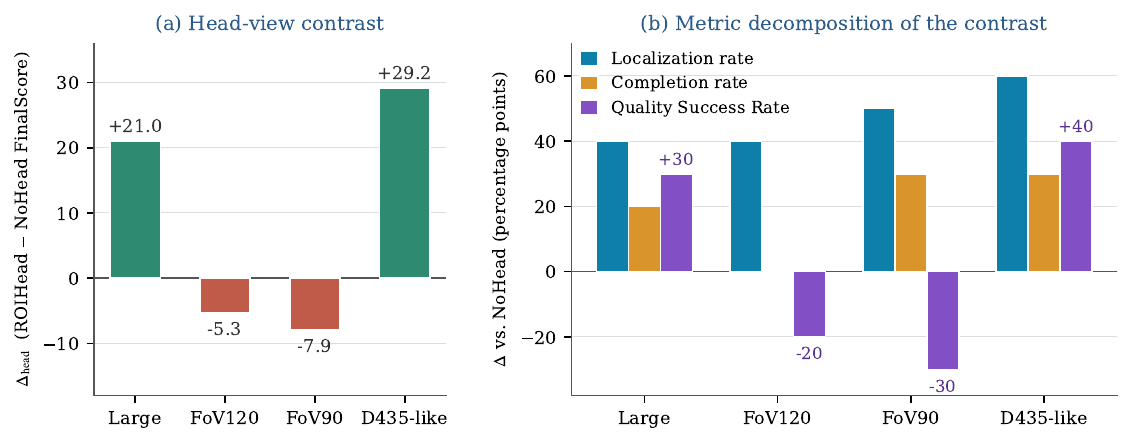}
\caption{Decomposition of the head-view contrast across wrist-FoV settings on Task B. (a) The FinalScore contrast $\Delta_{\mathrm{head}}$ (ROIHead $-$ NoHead) is positive under Large and D435-like but negative at the medium settings. (b) Metric-level changes explain this pattern: the localization rate improves under every FoV setting, whereas the Quality Success Rate flips sign at FoV120 and FoV90, where longer ROIHead trajectories violate the $T\leq 900$ high-quality gate (the completion-rate change at FoV120 is zero); see Table~\ref{tab:fov-scan} for the full per-setting values, including steps. Values are descriptive means over the fixed rollout slots.}
\label{fig:fov-delta}
\end{figure}

The FoV experiment uses a paired NoHead and ROIHead design under each FoV setting, as shown in Table~\ref{tab:wrist-fov-experiment}.
The corresponding contrast decomposition is visualized in Fig.~\ref{fig:fov-delta} together with the results below.
\begin{table}[!ht]\centering\small
\caption{Wrist-FoV experiment configurations.}
\label{tab:wrist-fov-experiment}
\begin{tabularx}{0.78\linewidth}{c c c Y}
\toprule
\textbf{FoV Setting} & \textbf{NoHead} & \textbf{ROIHead} & \textbf{Purpose}\\
\midrule
Large & F1 / G1 & F2 / G3 & Large-FoV baseline.\\
FoV120 & F5 & F6 & Medium-FoV comparison.\\
FoV90 & F7 & F8 & More restricted FoV comparison.\\
D435-like & F3 / G4 & F4 / G6 & Small-FoV comparison.\\
\bottomrule
\end{tabularx}
\end{table}

We now report the measured results of this scan; the head-view contrast $\Delta_{\mathrm{head}}$ follows Eq.~\ref{eq:head-gain}.
The main effect is not captured by monotonicity alone.
Across all four FoV settings, ROIHead improves the reported localization rate relative to NoHead, and the implied outcome contribution $\mathrm{FinalScore}-50q_{\mathrm{hq}}$ also increases when $q_{\mathrm{hq}}$ is expressed as a fraction.
The non-monotonicity in $\Delta_{\mathrm{head}}$ therefore comes mainly from the high-quality success gate rather than from coarse localization itself.
For Task B, high-quality success is an AND condition:
\begin{equation}
Q_{\mathrm{basket}} =
\mathbf{1}\{\operatorname{Comp}\,\wedge\,\operatorname{Loc}\,\wedge\,T\leq 900\,\wedge\,n_r\leq 1\},
\label{eq:basket-quality-gate}
\end{equation}
where $\operatorname{Comp}$ denotes Task B completion and $\operatorname{Loc}$ denotes passing the Task B localization criterion.
Thus, better localization only raises FinalScore when it also becomes stable completion with short trajectories and few retries.

For this reason, FinalScore should be read together with localization, completion, quality, and step count.
Table~\ref{tab:fov-scan} therefore keeps Completion Rate, Localization Rate, Quality Success Rate, and Average Steps.
This helps determine whether $\Delta_{\mathrm{head}}$ is associated with localization, quality, or changes in steps.
\begin{table}[!ht]\centering\scriptsize
\caption{FoV scan results on Task B. $\Delta_{\mathrm{head}}$ is the descriptive FinalScore difference between ROIHead and NoHead under the same FoV, computed from unrounded means before one-decimal display.}
\label{tab:fov-scan}
\begin{adjustbox}{width=\linewidth}
\begin{tabular}{c c c c c c c c}
\toprule
Wrist FoV & Head Type & FinalScore & Completion Rate & Localization Rate & Quality Success Rate & Avg. Steps & $\Delta_{\mathrm{head}}$\\
\midrule
Large & NoHead  & 68.2 & 60\% & 60\% & 50\% & 837.5 & --\\
Large & ROIHead & 89.2 & 80\% & 100\% & 80\% & 790.0 & +21.0\\
\midrule
FoV120 & NoHead  & 60.6 & 40\% & 50\% & 40\% & 885.0 & --\\
FoV120 & ROIHead & 55.3 & 40\% & 90\% & 20\% & 947.5 & -5.3\\
\midrule
FoV90 & NoHead  & 55.6 & 40\% & 50\% & 30\% & 935.0 & --\\
FoV90 & ROIHead & 47.8 & 70\% & 100\% & 0\% & 970.0 & -7.9\\
\midrule
D435-like & NoHead  & 42.2 & 20\% & 40\% & 10\% & 945.0 & --\\
D435-like & ROIHead & 71.4 & 50\% & 100\% & 50\% & 907.5 & +29.2\\
\bottomrule
\end{tabular}
\end{adjustbox}
\end{table}

Under the Large-FoV wrist setting, the positive Task B ROIHead contrast is best read as a contrast-specific complementary global-layout cue rather than compensation for broadly missing wrist information.
The wrist views still provide the main local gripper--object--contact evidence, while the ROI-cropped head view appears to add useful target and layout disambiguation in this setting.
This interpretation is consistent with the simultaneous improvements in localization ($+40$ percentage points, pp), completion ($+20$ pp), Quality Success Rate ($+30$ pp), and step count ($-47.5$ steps).
Here the head view does not replace the wrist views; it supplies a stable auxiliary cue that does not appear to conflict with local wrist-centered control in this contrast.

The FoV120 and FoV90 results show the opposite quality-gate behavior.
They illustrate that restricted wrist FoV can create an opportunity for the head view without guaranteeing a higher quality-adjusted score.
ROIHead improves localization by +40 pp and +50 pp, respectively, and the implied outcome contribution also increases.
However, the ROIHead average step counts are 947.5 for FoV120 and 970.0 for FoV90, both above the $T\leq 900$ high-quality threshold in Eq.~\ref{eq:basket-quality-gate}.
Consequently, better localization alone is not enough: a rollout can still lose the high-quality bonus if execution becomes slower, retry-heavy, or less stable at the final pose.
This is why Quality Success Rate drops by 20 pp for FoV120 and by 30 pp for FoV90, producing negative mean $\Delta_{\mathrm{head}}$ values despite improved localization.
A plausible mechanism is that these medium FoV settings sit between the two clearer regimes.
The wrist views still contain some local target evidence, but their global context is incomplete; ROIHead adds global context, yet the policy may trade off the wrist and head signals less consistently, which is associated with longer trajectories in these contrasts.

The D435-like setting is different.
Here the positive ROIHead gain is more consistent with compensation for missing or less stable wrist-view spatial evidence than with the complementary effect seen under Large FoV.
The wrist-only policy has less reliable global spatial evidence under the simulated restricted-FoV crop, so ROIHead provides concentrated layout information that is otherwise weak or absent.
The gain therefore appears not only in localization (+60 pp), but also in completion (+30 pp), Quality Success Rate (+40 pp), and step count ($-37.5$ steps).
Thus, D435-like ROIHead is best read as compensating for restricted wrist visibility in this Task B scan, whereas Large-FoV ROIHead is best read as a complementary global-layout cue.

Figure~\ref{fig:fov-delta} summarizes this decomposition: the FinalScore contrast is non-monotonic across FoV settings (panel a), while the underlying localization gain is uniformly positive and the quality-success change flips sign exactly at the medium settings (panel b).
The non-monotonic pattern should therefore be read as the result of the quality-gated score decomposition above, not as evidence that localization benefits are themselves non-monotonic.
The D435-like condition has the largest mean difference, while FoV120 and FoV90 show negative mean differences, although the descriptive bootstrap intervals overlap zero.
Overall, smaller wrist FoV can increase the potential value of a head view, but the realized quality-adjusted effect also depends on whether the policy can use the added view without longer, retry-heavy trajectories that violate the quality gate---not only on whether the added image provides dense, low-interference task-region information.

\subsection{Data-Scaling Experiment}

To study the relation between visual configuration and training data scale, we conduct a data-scaling experiment on Task A.
Task A includes several target objects and a long operation sequence.
It is therefore suitable for analyzing the learning efficiency of different visual configurations under different data scales.

The experiment uses three training-data scales:
\[
\{50,\ 200,\ 800\}\ \text{demonstrations}.
\]
The 50-demonstration dataset is included in the 200-demonstration dataset, and the 200-demonstration dataset is included in the 800-demonstration dataset.
This reduces the effect of data distribution differences.
All data scales share the same fixed budget of 60{,}000 optimization steps rather than a fixed number of epochs, so smaller subsets traverse their data more often; we keep the optimization budget fixed so that configurations are compared under equal training compute.
To separate the hypotheses of broader view coverage and higher task-region concentration, the data-scaling experiment keeps only three representative configurations: large-FoV dual wrist, small-FoV dual wrist, and small-FoV dual wrist plus ROIHead.
The settings are shown in Table~\ref{tab:scaling-configs}.

\begin{table}[!ht]\centering\small
\caption{Data-scaling experiment configurations.}
\label{tab:scaling-configs}
\begin{tabular}{c c c c}
\toprule
\textbf{Configuration} & \textbf{Wrist FoV} & \textbf{Head Type} & \textbf{Demonstrations}\\
\midrule
2W-Large & Large & NoHead & 50 / 200 / 800\\
2W-D435Crop & D435-like & NoHead & 50 / 200 / 800\\
3V-D435Crop-ROIHead & D435-like & ROIHead & 50 / 200 / 800\\
\bottomrule
\end{tabular}
\end{table}

\begin{figure}[!ht]\centering
\includegraphics[width=0.82\linewidth]{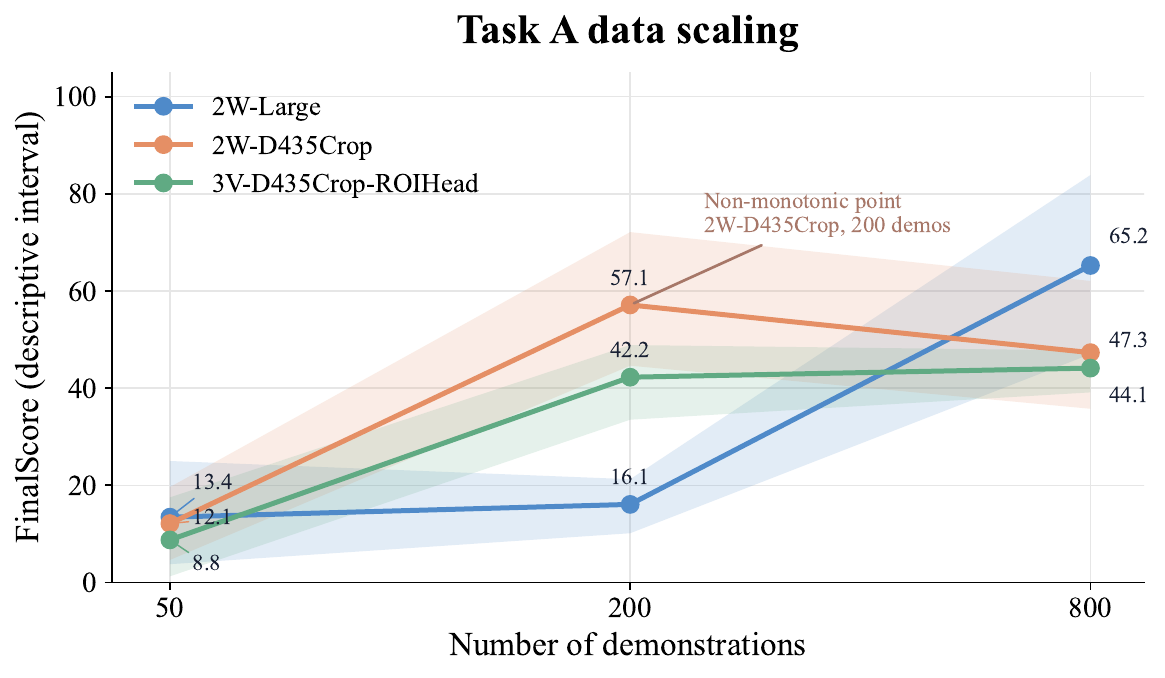}
\caption{Scaling curves. The x-axis is the number of demonstrations. The y-axis is FinalScore. The curves include 2W-Large, 2W-D435Crop, and 3V-D435Crop-ROIHead, with descriptive bootstrap intervals.}
\label{fig:scaling-results}
\end{figure}

This experiment asks whether large-FoV input needs more demonstrations because it includes more background and gives a lower proportion of task-relevant pixels.
We evaluate this trend through the learning curves reported in Fig.~\ref{fig:scaling-results}.

The results further show that the effect of FoV should not be evaluated without considering training data scale.
From the data-scale perspective, all three configurations improve substantially from 50 to 800 demonstrations.
However, the amount of data needed to reach a FinalScore threshold of 60 differs across configurations.
Gap$_{800-50}$ and $N_{60}$ in Table~\ref{tab:task-a-scaling} (Appendix~\ref{app:additional}) show this difference.
$N_{60}$ is estimated by piecewise linear interpolation over the observed 50/200/800 points when the score crosses 60; otherwise it is marked as not reached.

Large-FoV wrist cameras can cover a more complete operation area and spatial context.
This is their main advantage.
At the same time, large FoV also introduces more background and irrelevant regions.
It reduces the proportion of the target object under a fixed image resolution or fixed visual-token budget.
With small datasets, the model may not yet learn to focus reliably on key regions in the larger image.
The advantage of large FoV may therefore not appear immediately.

The experimental trend shows that the large-FoV configuration improves clearly as the number of demonstrations increases and is the only configuration whose mean FinalScore crosses the 60 threshold in the observed range.
This crossing is a statement about means: the descriptive intervals at these sample sizes are wide enough that the threshold is not separated from the other configurations, so the crossing indicates the direction of the trend rather than a resolved difference.
This suggests that large FoV may have a higher quality-adjusted performance ceiling, but may also need more data to become stable.
Both statements are based on mean FinalScore: the descriptive intervals remain wide, the intervals of 2W-Large and 2W-D435Crop still overlap at 800 demonstrations, and the interval of 2W-D435Crop at 200 demonstrations extends above 60, so the threshold crossing and the ceiling difference should be read as differences in means rather than separated results.
In contrast, small-FoV or ROIHead configurations can show stronger mid-scale completion but do not reach the same quality-adjusted score at 800 demonstrations in this dataset.

The curve shape also cautions against a simple scaling-law interpretation.
2W-Large rises clearly only at 800 demonstrations, which suggests that the advantage of large FoV depends more strongly on data scale under the quality-adjusted metric.
We observe a non-monotonic point at 200 demonstrations for 2W-D435Crop.
This point is consistent with high variance under the fixed evaluation schedule.
Figure~\ref{fig:scaling-results} shows the curves.

\subsection{Supplementary Diagnostic: Missing Target Visibility in Wrist Views}
\label{sec:supp-diagnostic}

The toy-transfer task is included as a supplementary boundary-case diagnostic.
The target toy starts inside either the left or the right bin, and the robot must grasp it from the source bin and transfer it into the other bin.
The initial target-selection stage contains a deliberate information gap: the head view observes both bins and distinguishes which bin contains the toy, whereas both wrist cameras face the outer bin walls and cannot observe the toy at the beginning of the task.
The purpose is therefore to inspect whether the available visual evidence around arm selection and target-bin choice changes when that auxiliary view is available.
The analysis focuses on this decision boundary rather than full task success.

\begin{figure}[H]\centering
\includegraphics[width=\linewidth]{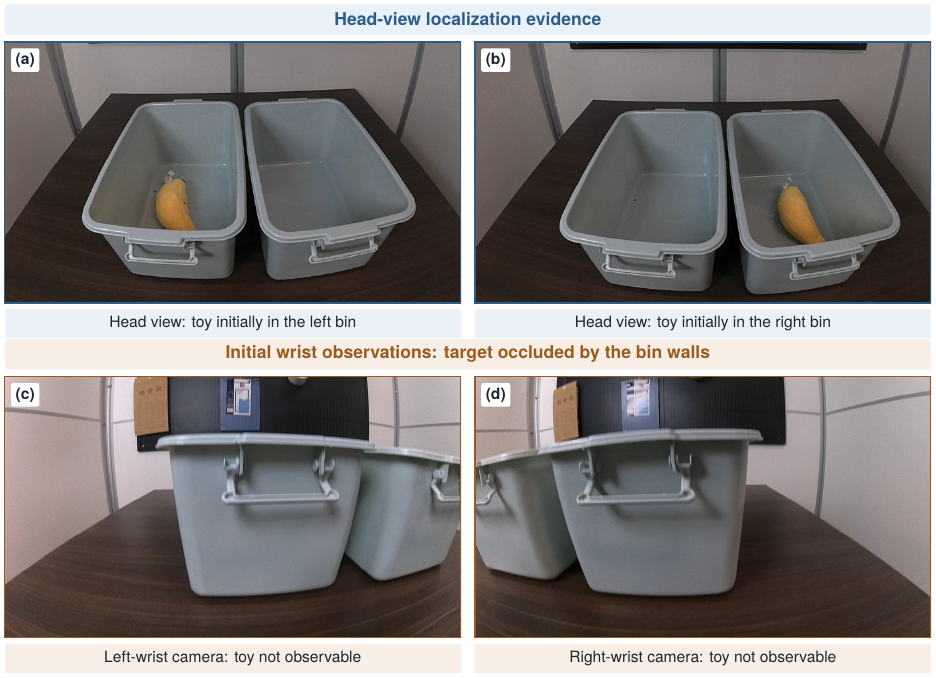}
\caption{Initial observability in the toy-transfer diagnostic. (a,b) The head view distinguishes whether the toy starts in the left or right bin. (c,d) The initial left- and right-wrist observations do not expose the target because of the bin walls. The panels document the visual information available at the initial decision stage; they are not rollout-success evidence.}
\label{fig:toy-transfer-observability}
\end{figure}

Figure~\ref{fig:toy-transfer-observability} shows the information available at the initial target-selection stage.
The head view distinguishes which bin contains the toy, whereas neither wrist view exposes the target behind the bin walls.
This is consistent with the conditional interpretation of this study: when wrist views cannot provide key target information, the head view can act as an auxiliary localization signal.

Figure~\ref{fig:toy-transfer-sensitivity-behavior} in Appendix~\ref{app:toy-diagnostic} provides complementary action-sensitivity and qualitative behavior examples.
In the rendered head-view maps, the dominant high-response region overlaps the toy-containing bin and changes sides with the visible target location.
The recorded choices are consistent with a left-bin preference in the shown wrist-only cases, but they do not quantify its frequency or cause.

This diagnostic remains a boundary-case analysis.
Because paired behavior records, head-view RISE coverage, and arm-selection statistics are less complete than in the main benchmark, we keep it as supplementary diagnostic evidence rather than a complete experiment.
The action-sensitivity overlays are used only as diagnostic interpretation and do not replace rollout success metrics.

\section{Analysis}
\label{sec:analysis}
This section analyzes the mechanisms behind the results in Section~\ref{sec:experiments}: why large-FoV wrist cameras already form a strong baseline, why the input form of the head view matters more than its presence, how individual views contribute to policy decisions, and how typical failures arise.

Table~\ref{tab:main-comparisons} maps each research question, in the order posed in the introduction, to the evidence that answers it.
\begin{table}[!ht]\centering\small
\caption{Research questions and the evidence that serves them. The rows follow the RQ order in the introduction; contrasts are computed from the G1--G6 results in Table~\ref{tab:main-results-overview}.}
\label{tab:main-comparisons}
\begin{tabularx}{0.95\linewidth}{c >{\centering\arraybackslash}p{0.30\linewidth} Y}
\toprule
\textbf{RQ} & \textbf{Evidence} & \textbf{What it answers}\\
\midrule
RQ1 & G2$-$G1,\ G3$-$G1 & How much gain RawHead or ROIHead adds when wrist FoV is large.\\
RQ2 & G5$-$G4,\ G6$-$G4;\ FoV scan (Table~\ref{tab:fov-scan});\ G1 vs.\ G6 & Whether restricted wrist FoV increases head-view value, and when that potential fails to raise the quality-adjusted score.\\
RQ3 & G3$-$G2,\ G6$-$G5 & Effect of ROI cropping relative to the raw head view under large and small wrist FoV.\\
RQ4 & Data-scaling experiment (50/200/800 demonstrations) & How demonstration count interacts with the visual-input configuration.\\
\bottomrule
\end{tabularx}
\end{table}

\subsection{Large-FoV Wrist Cameras Already Form a Strong Baseline}
Under the large-FoV dual wrist setting, the measured head-view effect is unstable.
The core comparisons are G2$-$G1 and G3$-$G1 in Table~\ref{tab:main-comparisons}.
In these comparisons, wrist FoV is fixed as Large, and only the presence and form of the head view are changed.

The results show that when the wrist cameras can already see the target object, grippers, and goal region, the head view provides limited additional information in this task family.
In some contrasts, adding the head view increases mean FinalScore or completion rate.
In other contrasts, the change is small, ambiguous, or negative.
It can also come with more steps or a lower Quality Success Rate.
This suggests that a head view is not necessary merely because it is available; its value depends on whether it adds dense task-region evidence rather than extra background or conflicting global context.

This pattern can be understood from the information source.
Wrist cameras move with the arms and approach the target during manipulation.
They continuously observe the relation among the gripper, object, and goal region.
In contrast, a fixed head view provides a more stable global image.
Its image changes less during many tabletop operations and may not contain the key contact details.
When wrist FoV is large enough, a broad head image is therefore more likely to be redundant than necessary, and it can lower quality-adjusted performance when the useful task region occupies a small fraction of the image.

This does not mean that the head view has no role under the large-FoV setting.
The more precise conclusion is that its effect is task-dependent when the large-FoV wrist views already cover the task-critical regions.
For example, Task B shows a positive Large-FoV ROIHead contrast, but this is contrast-specific evidence of a useful auxiliary layout cue rather than evidence that a head camera is generally required.
It should not be assumed that adding the head view always improves performance.

This task dependence is clear under the Large FoV setting.
If the wrist view is fixed and we only compare the addition of RawHead or ROIHead, G2$-$G1 and G3$-$G1 do not change in the same direction for FinalScore, Quality Success Rate, and Steps, as shown in Table~\ref{tab:large-fov-head-gain}.
\begin{table}[!ht]\centering\small
\caption{Head-view contrasts under the Large FoV setting. The comparisons are G2$-$G1 and G3$-$G1. Values show changes relative to the Large FoV NoHead baseline G1. FinalScore contrasts are computed from unrounded means before one-decimal display.}
\label{tab:large-fov-head-gain}
\begin{tabular}{c c c c c c}
\toprule
Task & Comparison & $\Delta$Completion Rate & $\Delta$FinalScore & $\Delta$Quality Success Rate & $\Delta$Steps\\
\midrule
Task A & G2$-$G1 RawHead & +40 pp & -16.0 & -40 pp & +70.0\\
Task A & G3$-$G1 ROIHead & +30 pp & -6.0 & -20 pp & +75.0\\
\midrule
Task B & G2$-$G1 RawHead & -10 pp & -1.1 & -10 pp & +67.5\\
Task B & G3$-$G1 ROIHead & +20 pp & +21.0 & +30 pp & -47.5\\
\midrule
Task C & G2$-$G1 RawHead & 0 pp & 0.0 & 0 pp & -5.0\\
Task C & G3$-$G1 ROIHead & +20 pp & +12.2 & +20 pp & -60.0\\
\bottomrule
\end{tabular}
\end{table}

The Task A rows in Table~\ref{tab:large-fov-head-gain} show why completion rate alone can be misleading.
G2 and G3 complete more often than G1 by +40 pp and +30 pp, respectively, but their mean FinalScore is lower.
This is not a contradiction in the scoring protocol.
Completion only records whether the task reaches the coarse task-completion condition.
FinalScore in Eq.~\ref{eq:quality-adjusted-score} also rewards high-quality completion, so a completed rollout can still receive a lower score when the final placement is inaccurate, the object state is unstable, the trajectory is long, or the execution requires repeated attempts.
In Task A, this appears as lower Quality Success Rate and longer average Steps for G2/G3 despite higher completion.
Table~\ref{tab:completion-quality-drop} in Appendix~\ref{app:scoring} summarizes common ways in which completion can fail to become high-quality completion across the task-specific quality gates.

\begin{figure}[!ht]\centering
\includegraphics[width=0.86\linewidth]{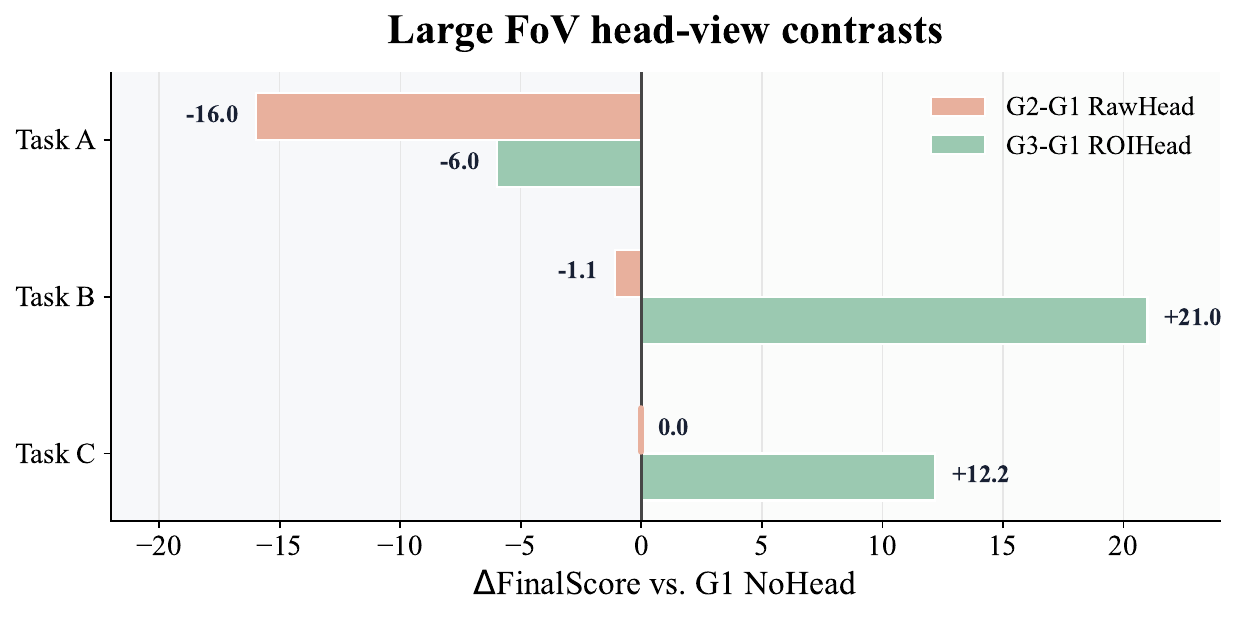}
\caption{Head-view contrasts under the Large FoV setting. Horizontal bars show the mean $\Delta$FinalScore for G2$-$G1 or G3$-$G1 relative to the G1 NoHead baseline. The zero line separates lower and higher mean scores relative to G1.}
\label{fig:large-fov-head-gain}
\end{figure}

A positive mean difference does not imply a stable improvement under the reported descriptive uncertainty.
Figure~\ref{fig:large-fov-head-gain} therefore shows only the sign and magnitude of the mean contrasts against G1, rather than presenting the contrast as a formal interval estimate.
The exceptions should be read as contrast-specific evidence, not as a general result that head-camera input improves the model.
Under the large-FoV setting, the head view is therefore better interpreted as a task-dependent auxiliary signal rather than a default performance requirement.
Figure~\ref{fig:large-fov-head-gain} shows this pattern.

\subsection{Head-View Form Matters More Than Head-View Presence}
The input form of the head view strongly affects its stability.
RawHead provides a wider global view, but it also includes more background, table boundaries, robot appearance, and task-irrelevant regions.
ROIHead restricts the head view to the task region.
This increases the proportion of the target object and operation area in the image.

The results show that RawHead does not always fail completely.
Its characteristic weakness is that completion and completion quality come apart: the widest gap between completion rate and Quality Success Rate in Table~\ref{tab:main-results-overview} is a RawHead cell, where Task A G2 completes 90\% of rollouts with a $0\%$ Quality Success Rate, and Task A G5 shows the same separation at 70\% and $0\%$.
This suggests that RawHead can supply enough coarse scene information for the policy to make progress while still failing to support precise localization and high-quality execution.
The separation between completion and quality is not exclusive to RawHead, so we read it as a characteristic RawHead failure mode rather than a discriminating signature.

ROIHead is more controlled as an input representation.
By removing much of the irrelevant background, ROIHead concentrates the head-view information around the target and operation region.
It may therefore provide a cleaner localization cue than RawHead when a head view is informative.
We do not claim that ROIHead is better than RawHead on every mean metric.
The more careful conclusion is that RawHead contrasts are unstable, while ROIHead is the cleaner head-view representation in the conditions where the third view matters.

This is also compatible with the observation that a larger head view is not always better.
For policy learning, the key factor is not view size alone.
The key factor is whether task-relevant information is concentrated, visible, and used consistently by the model.

The difference between RawHead and ROIHead should not be judged only by average score.
Completion quality and step count are also important.
Table~\ref{tab:raw-roi-contrast} therefore compares $\Delta$FinalScore, $\Delta$Quality Success Rate, and $\Delta$Steps.
This separates completing the task from completing it with high quality and fewer steps.
\begin{table}[!ht]\centering\scriptsize
\caption{RawHead versus ROIHead. The comparisons are G3$-$G2 and G6$-$G5. Values show ROIHead changes relative to RawHead. FinalScore contrasts are computed from unrounded means before one-decimal display.}
\label{tab:raw-roi-contrast}
\begin{adjustbox}{width=\linewidth}
\begin{tabular}{c c c c c c}
\toprule
Task & Comparison & $\Delta$FinalScore & $\Delta$Quality Success Rate & $\Delta$Steps & Brief Interpretation\\
\midrule
Task A & G3$-$G2 & +10.0 & +20 pp & +5.0 & ROI has a higher Quality Success Rate, but remains slower than RawHead.\\
Task A & G6$-$G5 & -0.9 & 0 pp & -35.0 & Mean score, quality rate, and step count are all close; this cell does not separate the two forms.\\
Task B & G3$-$G2 & +22.1 & +40 pp & -115.0 & ROI has a better mean contrast and fewer steps.\\
Task B & G6$-$G5 & +13.9 & +20 pp & -40.0 & ROIHead is the cleaner head-view form under restricted FoV.\\
Task C & G3$-$G2 & +12.2 & +20 pp & -55.0 & Under Large FoV, ROI is faster and has a higher Quality Success Rate.\\
Task C & G6$-$G5 & +21.4 & +20 pp & -20.0 & Under D435-like FoV, ROIHead gives the largest mean RawHead-to-ROIHead contrast.\\
\bottomrule
\end{tabular}
\end{adjustbox}
\end{table}

\begin{figure}[!ht]\centering
\begin{subfigure}[t]{0.48\linewidth}
  \centering
  \includegraphics[width=\linewidth]{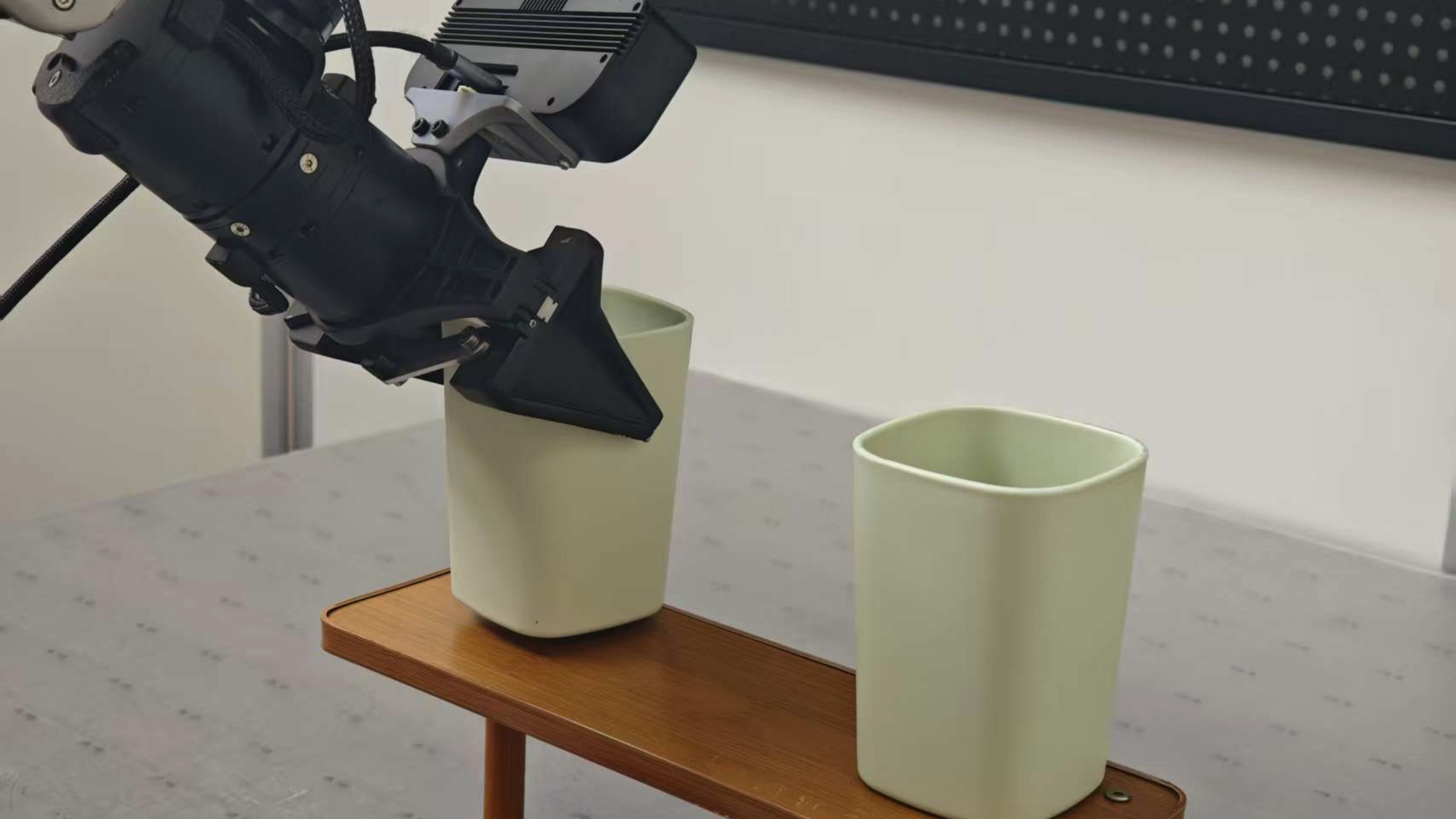}
  \caption{Task A: high-quality}
\end{subfigure}\hfill
\begin{subfigure}[t]{0.48\linewidth}
  \centering
  \includegraphics[width=\linewidth]{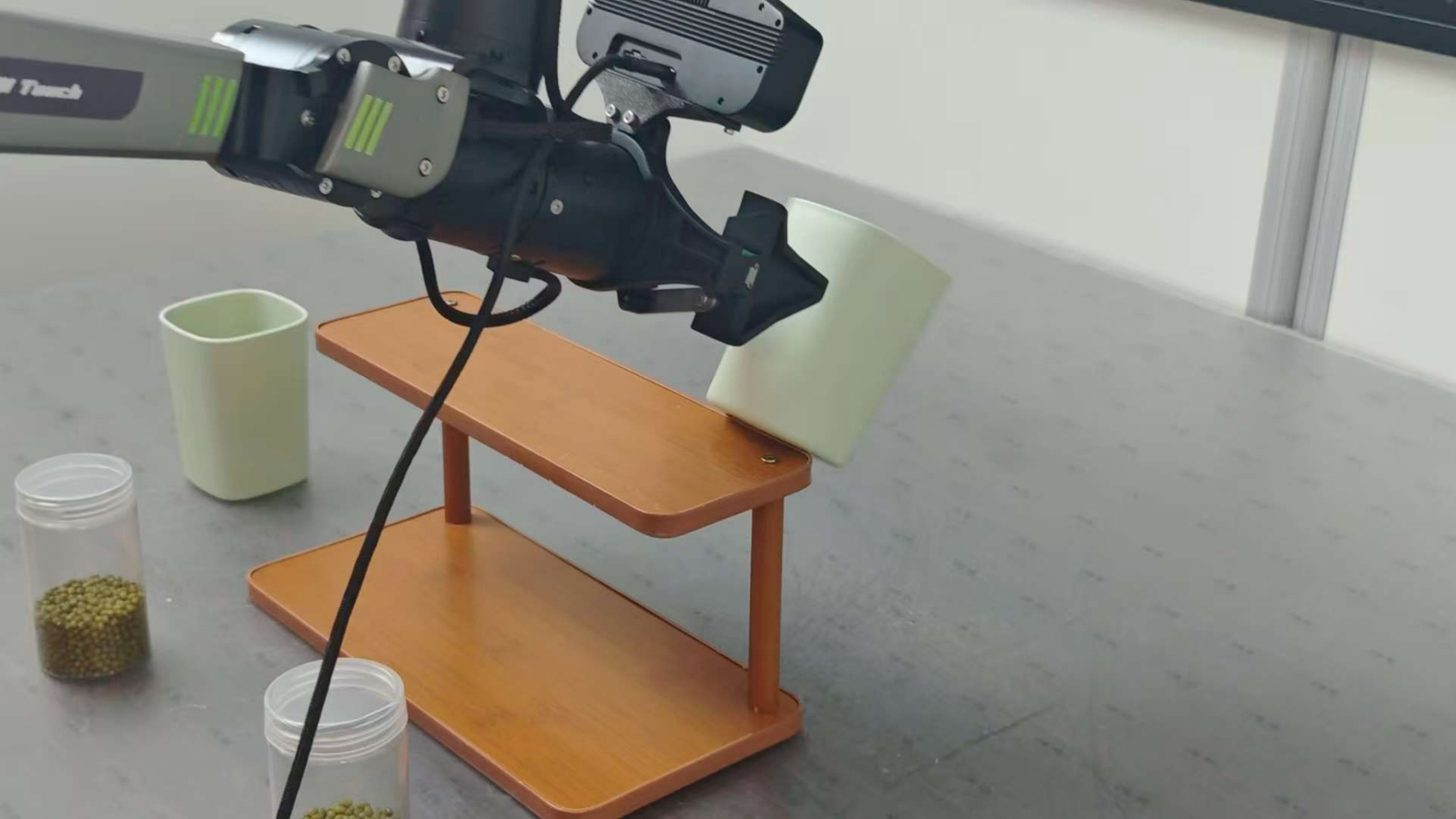}
  \caption{Task A: completion only}
\end{subfigure}\\[-1pt]
\begin{subfigure}[t]{0.48\linewidth}
  \centering
  \includegraphics[width=\linewidth]{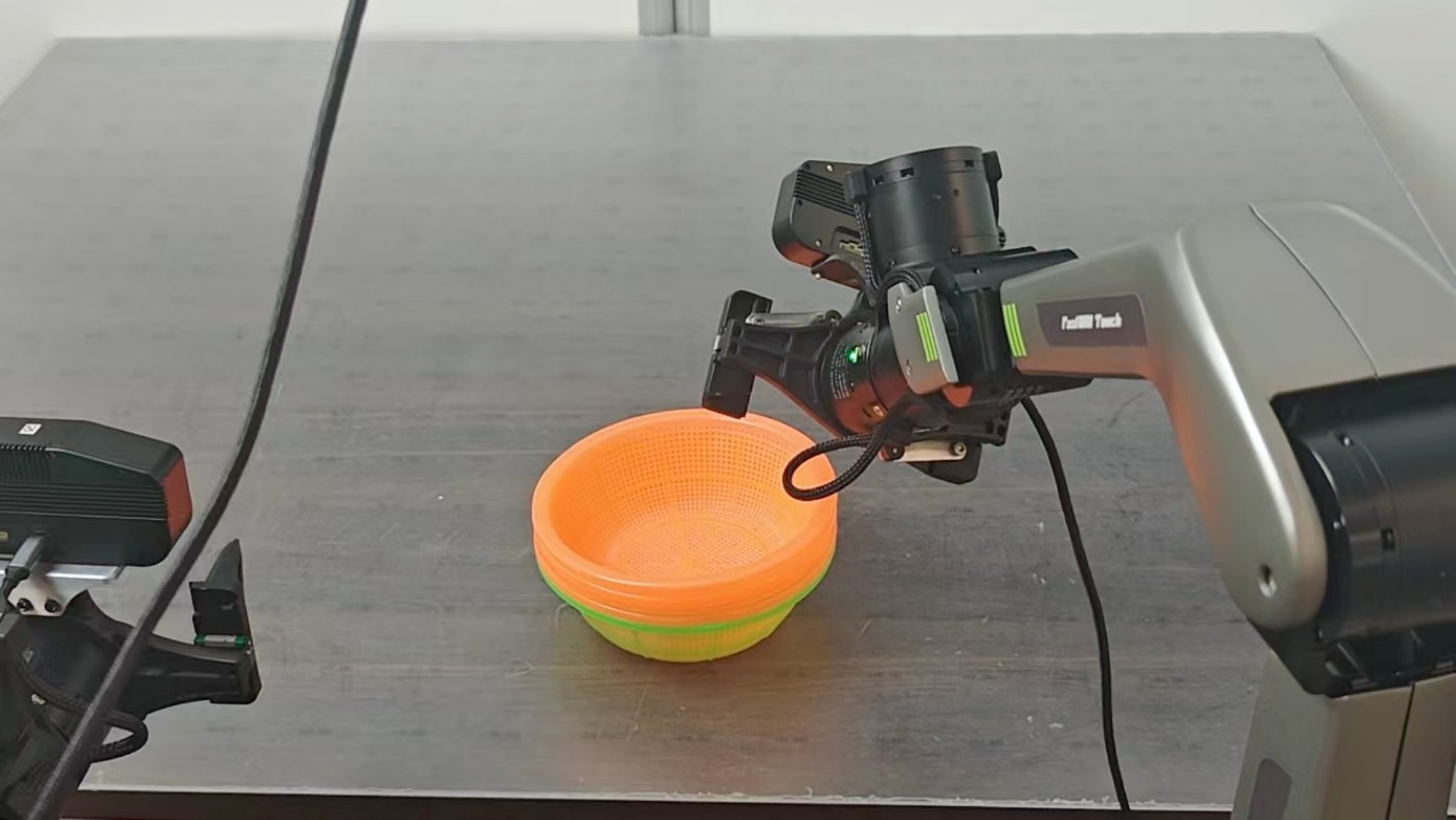}
  \caption{Task B: high-quality}
\end{subfigure}\hfill
\begin{subfigure}[t]{0.48\linewidth}
  \centering
  \includegraphics[width=\linewidth]{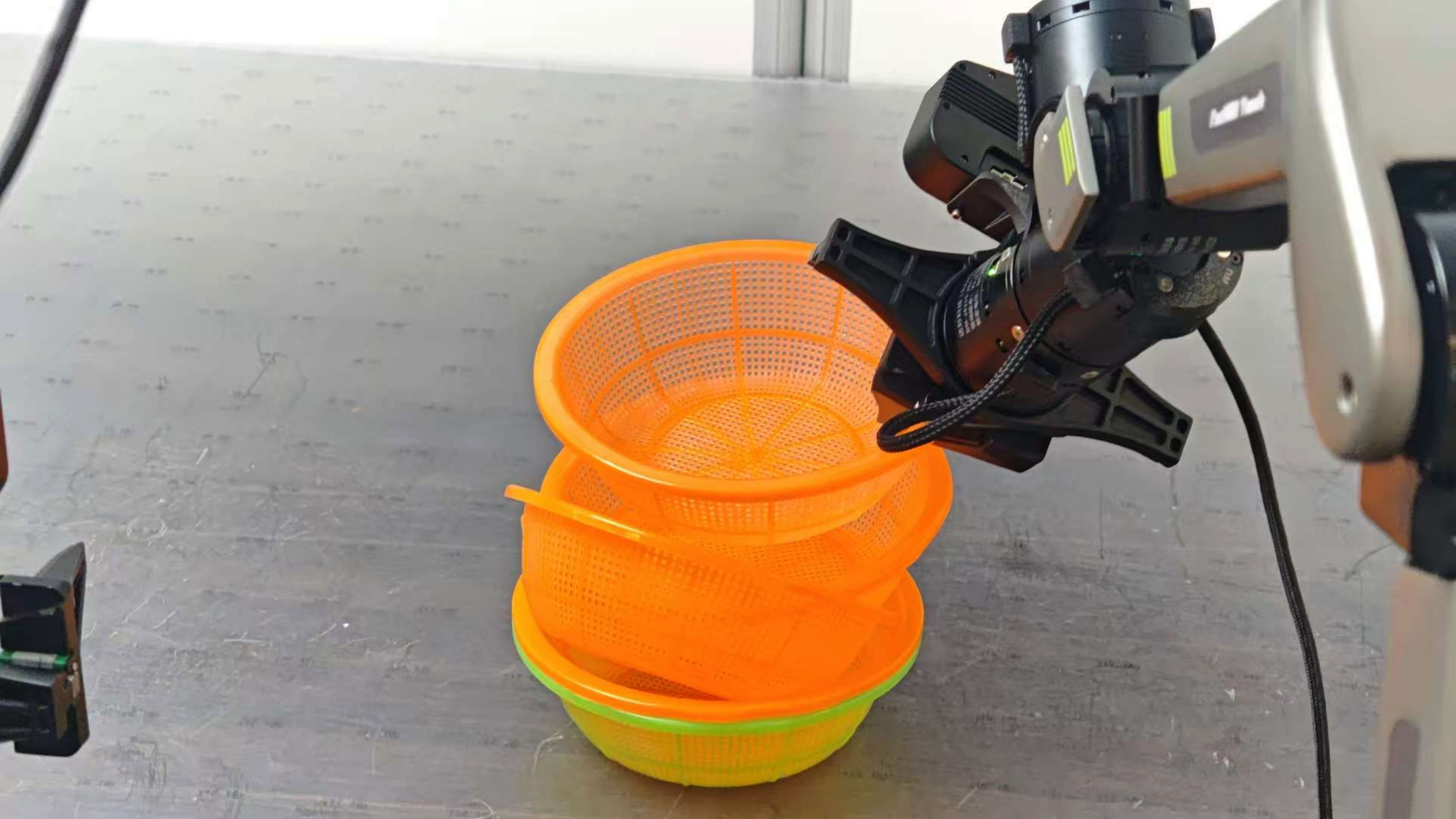}
  \caption{Task B: completion only}
\end{subfigure}
\caption{Representative real-robot final states illustrating the distinction between completion and high-quality completion. Completion-only examples can satisfy coarse task progress while still losing the high-quality bonus because of unstable final pose or poor alignment; long execution and repeated attempts are additional scoring-protocol failure modes summarized in Table~\ref{tab:completion-quality-drop}.}
\label{fig:real-quality-examples}
\end{figure}

Representative final-state images in Figure~\ref{fig:real-quality-examples} illustrate the same scoring distinction at the level of final pose and alignment.
They are qualitative examples of the completion-versus-quality rule, not additional statistical evidence.
The completion-only panels show that a rollout can make visible task progress or reach the coarse completion condition while still losing the high-quality bonus because the final state is unstable or poorly aligned.
Long execution and retry-heavy behavior are additional protocol-level causes summarized in Table~\ref{tab:completion-quality-drop}.
\FloatBarrier

This pattern can also be understood from the input images.
RawHead keeps large background and irrelevant regions, while ROIHead focuses the same head view on the task region.
As a result, ROIHead may make the head view easier to exploit when the third view contains relevant information, as shown in Figure~\ref{fig:raw-vs-roi-view} in Appendix~\ref{app:view-config}.

\begin{figure}[!ht]\centering
\includegraphics[width=0.86\linewidth]{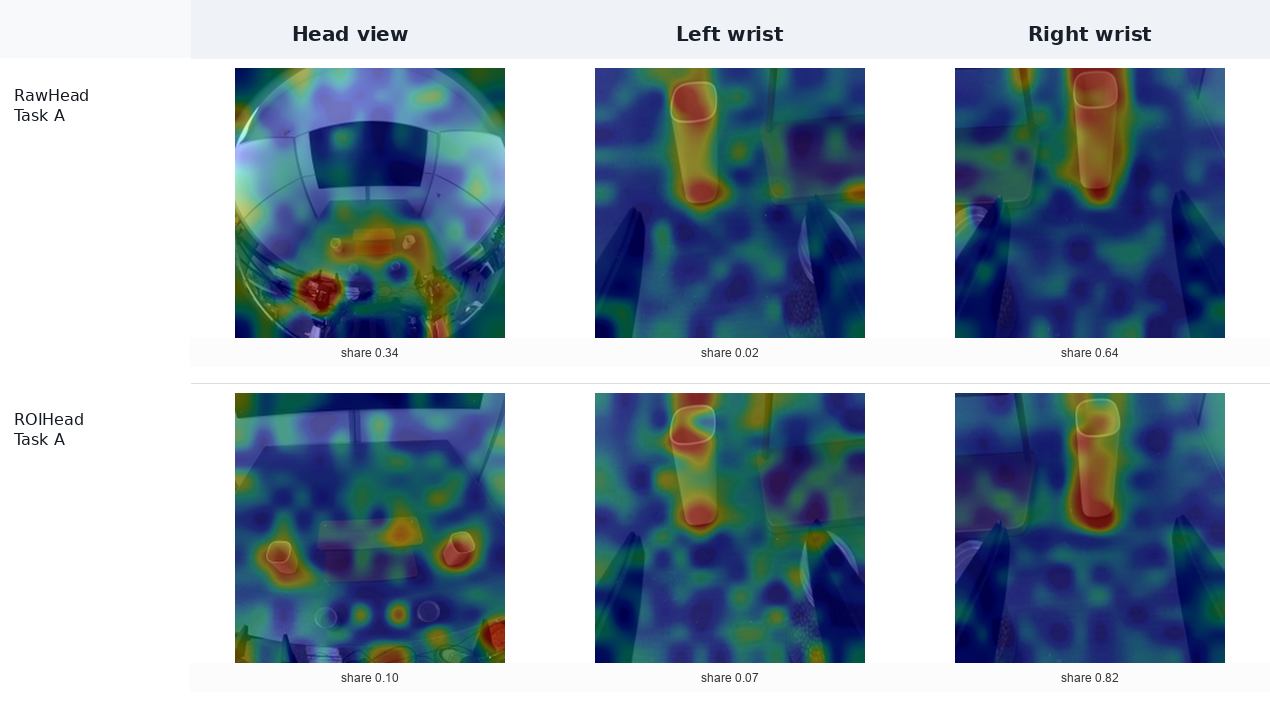}
\caption{RISE-style action-sensitivity example for RawHead and ROIHead on Task A. Heatmaps visualize $S_v(c)$ from Eq.~\ref{eq:rise-action-sensitivity}; action-share values report $q_v$ from Eq.~\ref{eq:view-share} within each selected step. The maps indicate local action sensitivity, not attention weights or rollout success.}
\label{fig:heatmap-raw-roi}
\end{figure}

RISE-style action-sensitivity maps illustrate a compatible interpretation.
Figure~\ref{fig:heatmap-raw-roi} shows a representative comparison between RawHead and ROIHead on Task A.
In this single-step diagnostic example, RawHead has a larger head-view share $q_v$ under Eq.~\ref{eq:view-share}, but its high-$S_v(c)$ regions can be spread over a broader scene region.
ROIHead restricts the head view to the task region.
In this example, the ROI-cropped head view behaves more like low-weight spatial context, while the wrist views show stronger local contact-region responses.
This observation does not imply that ROIHead is universally better than RawHead on all tasks or metrics.
It is compatible with the observed RawHead instability in the quantitative results.

\subsection{Visual Attribution: View Roles and Action Sensitivity}
To understand how different camera views contribute to policy decisions, we generate the RISE-style action-sensitivity maps defined in Eq.~\ref{eq:rise-action-sensitivity}.
Each map estimates how much the predicted action chunk changes when a local image region is hidden while the rest of the observation is unchanged.
The view share in Eq.~\ref{eq:view-share} summarizes the same diagnostic at the camera level for the selected step.
The maps and shares should not be interpreted as Transformer attention or rollout success.
Their role is to provide diagnostic evidence, not to replace real-robot rollout metrics.

\begin{figure}[!ht]\centering
\includegraphics[width=0.90\linewidth]{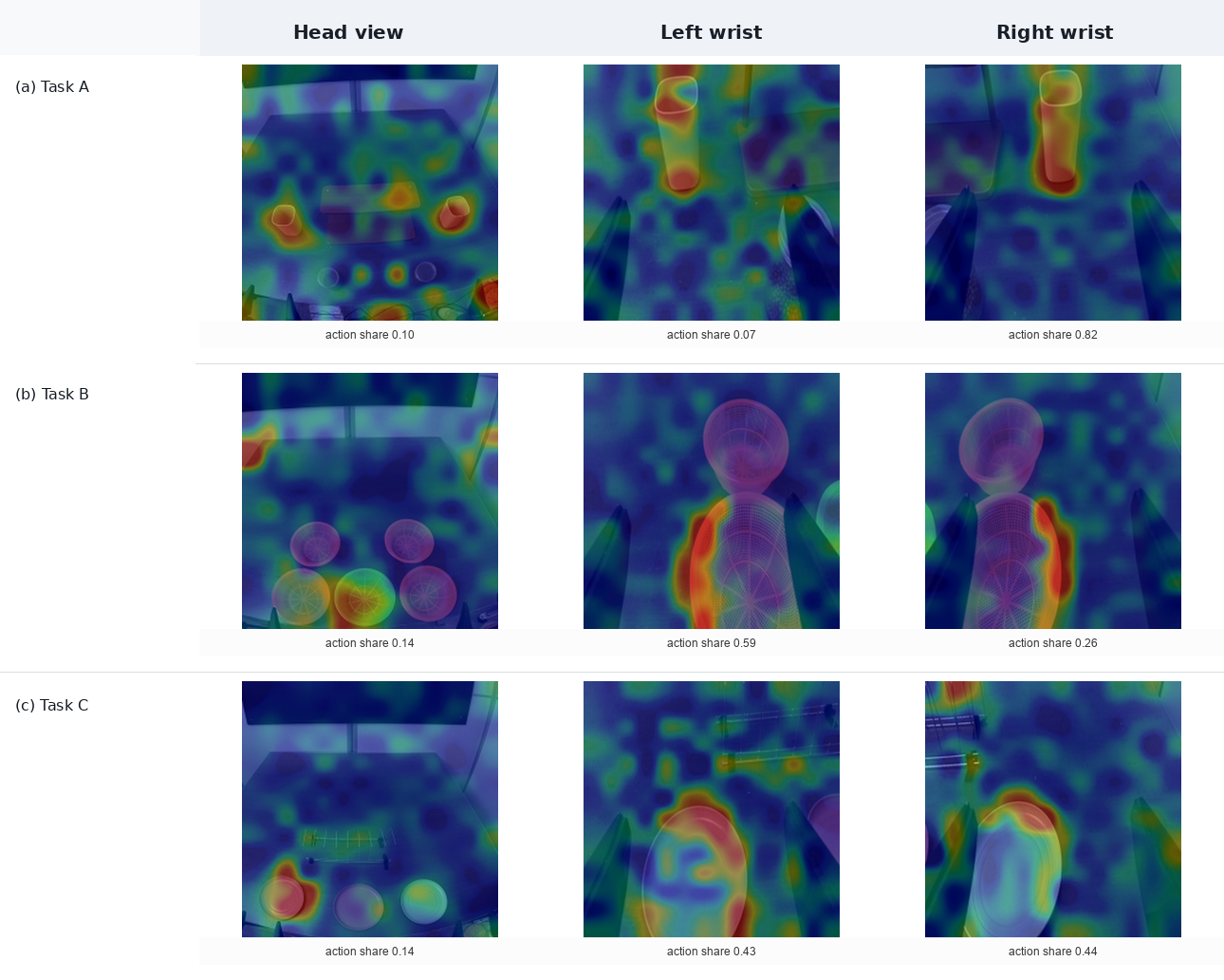}
\caption{RISE-style action-sensitivity examples across the three tasks. Heatmaps show $S_v(c)$, and view-level shares are $q_v$ within each selected step. Wrist views typically respond near grippers, objects, and contact regions, while the head view provides lower-weight global context. These examples are diagnostic illustrations of view roles and are not success-rate evidence.}
\label{fig:heatmap-view-roles}
\end{figure}

Across representative examples, high-response regions in the wrist views are concentrated around the gripper, object, and contact boundary.
This is compatible with wrist cameras acting as local-control inputs in these representative examples.
In contrast, responses in the head view more often appear around target containers, racks, or global placement regions.
In these representative examples, this suggests that the head view provides coarse localization and scene reference, rather than direct gripper-contact control.
Figure~\ref{fig:heatmap-view-roles} shows this division of roles across Task A/B/C.

Figure~\ref{fig:heatmap-nohead-roi} in Appendix~\ref{app:additional} further compares NoHead and ROIHead under the same restricted-wrist-FoV family: the NoHead panels provide a two-wrist reference, while the ROIHead panels show the additional third view.
In the selected ROIHead single-step examples for Task A/B/C, the head-view action shares are about 0.10, 0.14, and 0.14.
Because $q_v$ is normalized over valid views within a selected step, the remaining share in those examples is assigned to the wrist views.
Thus, in these examples, the head view provides weaker but spatially meaningful auxiliary signals.
This is compatible with the quantitative results in these tabletop bimanual tasks: the head view behaves more like an auxiliary localization signal than a replacement for wrist-centered local control.

\subsection{Failure Mode Analysis}
To avoid explaining model behavior only through mean scores, we also summarize typical failure modes from scoring records and qualitative annotations of recorded rollouts.
These cases do not change the quantitative conclusions.
They provide diagnostic interpretations for different view configurations.
The failure-mode column is based on rollout scoring and qualitative rollout notes, while the action-sensitivity statements are based only on representative RISE-style examples.

\begin{table}[!ht]\centering\scriptsize
\caption{Typical failure modes and diagnostic interpretations. Scoring signatures come from rollout metrics; action-sensitivity statements come from representative heatmap examples and are diagnostic rather than causal evidence.}
\label{tab:failure-modes}
\begin{adjustbox}{width=\linewidth}
\begin{tabular}{p{0.13\linewidth} p{0.27\linewidth} p{0.32\linewidth} p{0.24\linewidth}}
\toprule
\textbf{Configuration} & \textbf{Typical Failure Mode} & \textbf{Diagnostic Interpretation} & \textbf{Observed Signature}\\
\midrule
RawHead &
The task can progress or even finish, but the final quality is low, steps are long, or retries increase. &
RawHead contains more background and irrelevant regions. The target occupies a smaller image area. The policy may get coarse scene context but unstable fine localization. &
Completion rate is not always low while Quality Success Rate is: on Task A, G2 completes 90\% of rollouts and G5 70\%, both with a 0\% Quality Success Rate; some tasks also show increased Steps.\\
\midrule
D435-like NoHead &
The target, goal, or gripper-goal relation becomes partly invisible at critical moments and is associated with localization or placement errors. &
Cropping increases local pixel density but removes global geometry. Once the target or goal leaves the wrist image, the policy has little compensating information. &
In Task B/C, G4 is more likely to show low FinalScore or low localization rate than the Large Wrist baseline.\\
\midrule
Large Wrist NoHead &
The policy often completes the task, but global localization can still be unstable in harder layouts or dispersed-object states. &
Large-FoV wrist views cover broad local context, but they still move with the arms. Distant goals and cross-object spatial relations are not always stable. &
G1 is a strong baseline, but Task A/C still contain low-quality completions or long trajectories.\\
\midrule
ROIHead &
Global localization can improve, but wrist-centered local control is not fully replaced. Failures can still occur during contact, grasping, or insertion. &
ROIHead provides concentrated spatial context. However, gripper-object-contact cues are still often supplied by the wrist views. &
In representative action-sensitivity examples, wrist views have larger local contact-region responses. G6 has a higher mean FinalScore than G4 on Task B and Task C but not on Task A, and its Quality Success Rate does not always reach Large Wrist.\\
\midrule
Restricted wrist FoV (any head type) &
The policy holds its initial pose for an extended period before the first motion, and in a small number of rollouts one arm never executes a grasp, so the objects assigned to that arm are never attempted. &
With the wrist images cropped, the evidence that triggers the first approach is weaker at the initial pose, where the arm is far from every target. Delayed or absent initiation consumes the step budget without producing task progress. &
On Task A, one rollout in each of G5 and G6 records only two grasps; both are at the highest step count observed in their configuration ($1500$), and the corresponding recordings show a single arm idle for the whole episode.\\
\bottomrule
\end{tabular}
\end{adjustbox}
\end{table}

Task A under the restricted wrist FoV shows why the binary completion rate and the outcome score can diverge.
G6 completes only 30\% of its rollouts against 70\% for G5, yet the two configurations differ by about one point in mean FinalScore ($44.1$ versus $45.1$).
The rollout records locate the difference at the end of the episode rather than at its start: in six of the seven G6 non-completions all four objects were grasped, and the rollout failed only in the final state---four placed all four objects but left one of them unstable, and two left one object outside its assigned position.
Because completion requires all four objects to be simultaneously grasped, correctly placed, and stable, such a rollout is recorded as a non-completion even though its outcome score $P$ reaches $92.5$ out of $100$.
The completion rate therefore understates how close the restricted-FoV ROIHead configuration comes to finishing the task, while the outcome score, which awards partial credit per object, records the same difference as small.
This is the mirror image of the Task A large-FoV pattern discussed above, where completion is high and quality is low.

Step counts increase with wrist-FoV restriction rather than with head-view configuration.
Averaged over the three head-view forms, the D435-like group needs more low-level control steps than the Large group on every task: $1188$ versus $1023$ on Task A ($+16\%$), $933$ versus $844$ on Task B ($+11\%$), and $1288$ versus $1078$ on Task C ($+20\%$).
The increase is also present without any head view---on Task A, the wrist-only G4 has the joint-highest step count---so it cannot be attributed to the presence of an additional view.
This is consistent with the information-density reading of the main results: once the wrist images no longer contain the target and the goal region at the same time, the policy spends additional steps initiating, re-approaching, and re-locating before it commits to contact.

These failure cases reinforce the main claim of the report.
The key variable is not the number of cameras, but the visibility and density of task-relevant information in the input images.
RawHead failures are consistent with the view being too broad and background-heavy.
D435-like NoHead failures are consistent with critical regions being cropped away.
ROIHead can add global-localization evidence in some restricted-view cases, while representative sensitivity examples still show stronger wrist-view responses around local contact regions.

\subsection{Answers to the Research Questions}
\label{sec:rq-answers}
We close the analysis by answering the four research questions posed in the introduction.

\begin{description}
  \item[RQ1: how much additional gain does a head view provide under large-FoV wrists?]
  Little on average, and not consistently across tasks.
  Under the Large wrist setting, the head-view contrasts do not move FinalScore, Quality Success Rate, and Steps in a consistent direction (Table~\ref{tab:large-fov-head-gain}): RawHead lowers mean FinalScore on Task A ($-16.0$) and Task B ($-1.1$), and ROIHead helps on Task B ($+21.0$) and Task C ($+12.2$) but not on Task A ($-6.0$).
  A head view is therefore a task-dependent auxiliary signal rather than a default requirement when wrist coverage is already large.

  \item[RQ2: does restricted wrist FoV increase the potential value of a head view, and when does that potential fail?]
  Restricting wrist FoV does increase the opportunity: ROIHead improves the localization rate under every FoV setting in the Task B scan (Table~\ref{tab:fov-scan}).
  However, the potential fails to become a higher quality-adjusted score at the medium settings (FoV120 $-5.3$, FoV90 $-7.9$) because the added view is associated with longer trajectories that violate the $T\leq 900$ quality gate, whereas it converts into a large gain under D435-like ($+29.2$); the separate positive contrast under Large ($+21.0$) reflects a complementary global-layout cue rather than restriction-induced head-view value.

  \item[RQ3: does an ROI-cropped head view provide denser, lower-interference information than a raw head view?]
  Yes, in the conditions where the head view matters.
  The ROI-versus-raw contrasts are positive on five of six task--FoV combinations (Table~\ref{tab:raw-roi-contrast}); representative action-sensitivity examples are compatible with this reading.
  This contrast does not depend on how FinalScore weights its two terms: counted directly on the component metrics, ROIHead attains the higher Quality Success Rate on five of the same six combinations and the lower step count on five of six, so the ordering is reproduced by metrics that involve no score weighting.

  \item[RQ4: how does the number of demonstrations influence the effect of visual-input configuration?]
  The comparison is data-scale-dependent.
  Under a fixed optimization budget, the large-FoV wrist-only configuration is the only one whose mean FinalScore crosses the 60 threshold, and it does so only at 800 demonstrations (Table~\ref{tab:task-a-scaling}), whereas the restricted-FoV configurations reach stronger mid-scale completion but do not reach the same quality-adjusted score at 800 demonstrations, which suggests a lower ceiling in the observed range; FoV conclusions should therefore not be drawn at a single data scale.
\end{description}

\section{Discussion and Limitations}

This study provides controlled evidence for camera-configuration decisions in tabletop manipulation.
It does not establish a universal law for multi-view robot manipulation.
In our setting, large-FoV wrist cameras already cover the grippers, objects, and goal regions.
This is compatible with the head view often behaving as an auxiliary localization signal rather than the dominant source of action sensitivity.
The conclusion should not be directly extended to mobile manipulation or broader scene-level tasks.
In mobile manipulation, navigation, long-range target search, re-localization after occlusion, and semantic scene understanding may make head or exocentric views essential.

The study has several limitations.
First, each configuration is evaluated on a small fixed set of real-robot rollouts, and the initial-layout set is fixed.
The statistics should therefore be read as indicative trends rather than conclusive evidence.
Second, the D435-like condition is simulated by cropping wrist images.
It does not reproduce the full imaging properties of a physical Intel RealSense D435 camera, including distortion, noise, exposure, and calibration effects.
Thus, the D435-like results should be interpreted as evidence about restricted FoV, not as a general claim about a specific hardware model.

Third, the RISE maps measure the normalized action changes in Eq.~\ref{eq:action-sensitivity-delta} under random image masking.
They indicate local action sensitivity as defined by Eq.~\ref{eq:rise-action-sensitivity}.
They are not Transformer attention maps, and they cannot be used as direct success-rate evidence.
Fourth, the data-scaling experiment fixes the optimization budget at 60{,}000 steps across demonstration counts, so smaller subsets are trained for many more epochs; the observed trends therefore reflect data scale under a fixed compute budget rather than a matched epoch count. The experiment also contains a non-monotonic point at 200 demonstrations for 2W-D435Crop.
For this reason, we do not present the trend as a strict scaling law.
We therefore treat this result as consistent with high variance under the fixed evaluation schedule, and only argue that FoV and data scale appear coupled in the current setting.
Finally, the toy-transfer task is a supplementary diagnostic.
It illustrates a setting in which the initial wrist views omit the target location while the head view reveals it, but it still requires more complete behavior records and attribution results.

\section{Conclusion}

This study examines whether a head view is still necessary for tabletop bimanual manipulation.
The results show that large-FoV dual wrist cameras form a strong baseline in this task family.
Wrist views stay close to the interaction process and provide continuous information about grippers, objects, contact regions, and many goal regions.
Under this condition, the head view is not a default requirement, and its measured effect remains task- and input-form-dependent under the reported descriptive uncertainty.

When wrist FoV is restricted, the target is occluded, or the task depends more strongly on global localization, a head view can provide useful auxiliary spatial evidence.
This opportunity is not realized monotonically: across the Task B wrist-FoV scan the ROIHead-versus-NoHead contrast is positive at the widest and at the narrowest wrist setting but negative at both intermediate settings, so restricting wrist FoV creates the opportunity for a head view without guaranteeing a gain.
The results also show that this benefit is not determined by view count alone: a broad raw head image can be low-density and high-interference, while an ROI-cropped head view is more likely to provide concentrated task-region context.
These comparisons are in addition data-scale-dependent: in the Task A scaling study the ordering of the wrist-FoV settings changes between 200 and 800 demonstrations, so a FoV conclusion drawn at one data scale need not transfer to another.
The practical design criterion is therefore whether the added view supplies high-density, low-interference information that the wrist views do not already provide and that the policy can use without degrading execution quality.

\bibliographystyle{IEEEtran}
\bibliography{references}

\appendix
\renewcommand{\thetable}{\thesection\arabic{table}}
\renewcommand{\thefigure}{\thesection\arabic{figure}}

\section{Scoring Details}
\label{app:scoring}
\setcounter{table}{0}\setcounter{figure}{0}
This appendix collects the detailed scoring criteria.
Table~\ref{tab:scoring-protocol} lists the completion and high-quality definitions for each task, and Table~\ref{tab:completion-quality-drop} summarizes typical completed-but-low-quality outcomes that reduce the high-quality bonus.

\begin{table}[!ht]\centering\scriptsize
\caption{Completion and high-quality criteria. For each task, the Completion Definition specifies when a rollout counts as task completion in the outcome score $P_{\tau}$, and the High-Quality Definition specifies the condition for the high-quality indicator $Q_{\tau}$ in Eq.~\ref{eq:quality-adjusted-score}. A key failure denotes an execution error that requires an evident recovery or repeated attempt during the rollout.}
\label{tab:scoring-protocol}
\begin{adjustbox}{width=\linewidth}
\begin{tabular}{p{0.15\linewidth} p{0.43\linewidth} p{0.36\linewidth}}
\toprule
\textbf{Task} & \textbf{Completion Definition} & \textbf{High-Quality Definition}\\
\midrule
Task A: object organization &
All four target objects are grasped, placed in the target region, and stable at the end of the rollout. &
Completion with no key failure and $T\leq 950$.\\
\midrule
Task B: basket nesting &
All four baskets are grasped, inserted, and stably nested into the target basket with acceptable final poses. &
Completion with localization passed, $T\leq 900$, and $n_r\leq 1$.\\
\midrule
Task C: plate placement &
All plates are grasped and inserted into the rack with an acceptable final state. &
Full completion, at least two correct first-localization decisions, $n_r\leq 1$, and $T\leq 1100$.\\
\bottomrule
\end{tabular}
\end{adjustbox}
\end{table}

\begin{table}[!ht]\centering\scriptsize
\caption{Completion-only outcomes that can reduce the high-quality bonus. The table clarifies why a configuration can have higher completion rate but lower FinalScore under the quality-adjusted metric.}
\label{tab:completion-quality-drop}
\begin{adjustbox}{width=\linewidth}
\begin{tabular}{p{0.22\linewidth} p{0.38\linewidth} p{0.34\linewidth}}
\toprule
\textbf{Case} & \textbf{Typical completed-but-low-quality state} & \textbf{Effect on metrics}\\
\midrule
Inaccurate final placement &
The object, basket, or plate reaches the target region or rack, but its pose, alignment, nesting, or insertion state is not clean enough for the high-quality criterion. &
Completion can be counted, while $q_{\mathrm{hq}}$ and the high-quality bonus remain low.\\
\midrule
Unstable final state &
The task appears complete at the coarse level, but the final object state is tilted, not stably seated, or close to falling out of the target region. &
FinalScore can decrease relative to a lower-completion but cleaner baseline.\\
\midrule
Long trajectory &
The robot eventually finishes the task, but the rollout exceeds the task-specific step threshold in Table~\ref{tab:scoring-protocol}. &
Steps increase and the high-quality indicator can be lost even when task progress is high.\\
\midrule
Retry-heavy execution &
The robot completes after repeated re-grasping, re-alignment, or insertion attempts. &
The rollout may satisfy completion but fail the retry or stability part of the high-quality gate.\\
\bottomrule
\end{tabular}
\end{adjustbox}
\end{table}

\section{View-Configuration Details}
\label{app:view-config}
\setcounter{table}{0}\setcounter{figure}{0}
This appendix details the visual-input configurations.
Table~\ref{tab:view-factors} defines the wrist-camera FoV settings and head-view types, Table~\ref{tab:view-visibility} reports the qualitative view-visibility audit, and Figure~\ref{fig:raw-vs-roi-view} shows a representative RawHead and ROIHead frame pair.

\begin{table}[!ht]\centering\small
\caption{The two controlled visual-input factors. Panel (a) lists the wrist-camera FoV settings, which progressively restrict the visible workspace from the original UMI-style large-FoV view to a narrow D435-like crop. Panel (b) lists the head-view types, which determine whether the head camera is provided to the policy and in which input form.}
\label{tab:view-factors}
\textbf{(a) Wrist-camera FoV settings}\\[0.3em]
\begin{tabularx}{0.78\linewidth}{c Y}
\toprule
\textbf{Wrist FoV} & \textbf{Description}\\
\midrule
Large & Original UMI-style large-FoV dual wrist setup, used as a strong baseline.\\
FoV120 & Medium-FoV setting.\\
FoV90 & More restricted medium-FoV setting.\\
D435-like & Wrist images are cropped to about $68^\circ$ FoV to simulate restricted wrist vision.\\
\bottomrule
\end{tabularx}\\[0.9em]
\textbf{(b) Head-view types}\\[0.3em]
\begin{tabularx}{0.78\linewidth}{c Y}
\toprule
\textbf{Head Type} & \textbf{Description}\\
\midrule
NoHead & Uses only the two wrist-camera inputs.\\
RawHead & Adds the uncropped raw head view.\\
ROIHead & Adds a head view cropped around the task region.\\
\bottomrule
\end{tabularx}
\end{table}

\begin{table}[!ht]\centering\scriptsize
\caption{View-visibility audit on representative frames. The table uses qualitative relative bins to describe the density of task-relevant information.}
\label{tab:view-visibility}
\begin{adjustbox}{width=\linewidth}
\begin{tabular}{c c c c c c}
\toprule
\textbf{View} & \textbf{Object Visible} & \textbf{Gripper Visible} & \textbf{Goal Visible} & \textbf{Target Pixel Ratio} & \textbf{Background Ratio}\\
\midrule
Large Wrist & High & High & Medium--High & Medium--High & Low--Medium\\
D435-like Wrist & Partial--High & High & Low--Partial & High when visible & Low--Medium\\
RawHead & High & Partial & High & Low & High\\
ROIHead & High & Partial & High & Medium--High & Medium\\
\bottomrule
\end{tabular}
\end{adjustbox}
\end{table}

\begin{figure}[!ht]\centering
\includegraphics[width=0.66\linewidth]{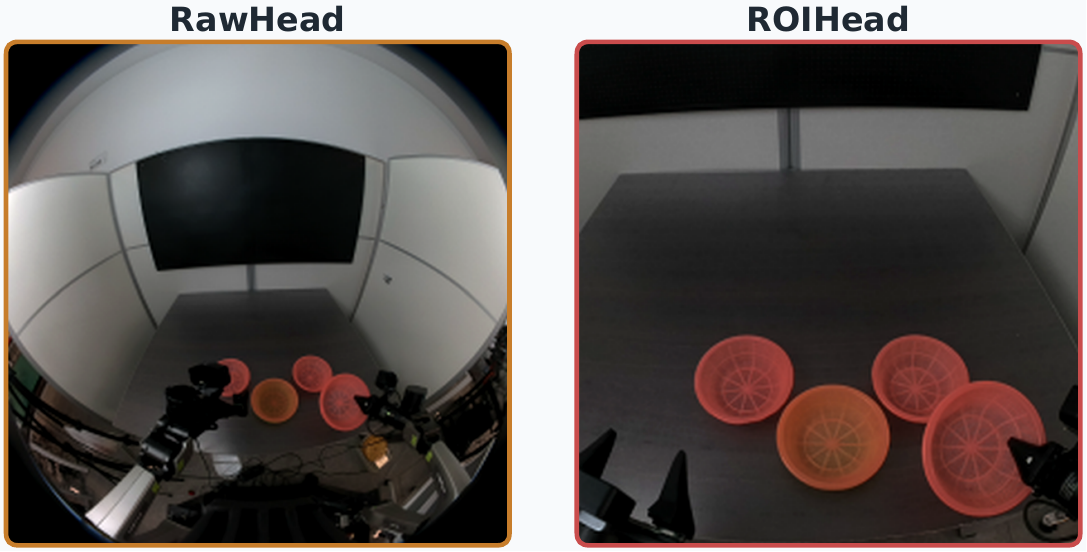}
\caption{Representative RawHead and ROIHead policy inputs from the basket-nesting experiments. RawHead retains broad scene context with lower task-region concentration, whereas ROIHead concentrates the task-relevant region. The panels come from separate rollouts and illustrate view geometry rather than a frame-matched crop comparison.}
\label{fig:raw-vs-roi-view}
\end{figure}

\section{Additional Results and Attribution Examples}
\label{app:additional}
\setcounter{table}{0}\setcounter{figure}{0}
This appendix provides additional quantitative and diagnostic material.
Table~\ref{tab:task-a-scaling} reports the full Task A data-scaling results, and Figure~\ref{fig:heatmap-nohead-roi} compares action-sensitivity maps between NoHead and ROIHead under the restricted wrist FoV.

\begin{table}[!ht]\centering\scriptsize
\caption{Scaling results on Task A. Gap$_{800-50}$ is the FinalScore difference between 800 and 50 demonstrations.}
\label{tab:task-a-scaling}
\begin{adjustbox}{width=\linewidth}
\begin{tabular}{c c c c c c c}
\toprule
Configuration & Demos & FinalScore & Completion Rate & Quality Success Rate & Gap$_{800-50}$ & $N_{60}$\\
\midrule
\multirow{3}{*}{2W-Large}
& 50  & 13.4 & 10\% & 0\%  & \multirow{3}{*}{+51.8} & \multirow{3}{*}{736}\\
& 200 & 16.1 & 0\%  & 0\%  &  & \\
& 800 & 65.2 & 50\% & 40\% &  & \\
\midrule
\multirow{3}{*}{2W-D435Crop}
& 50  & 12.1 & 0\%  & 0\%  & \multirow{3}{*}{+35.2} & \multirow{3}{*}{--}\\
& 200 & 57.1 & 80\% & 20\% &  & \\
& 800 & 47.3 & 60\% & 10\% &  & \\
\midrule
\multirow{3}{*}{3V-D435Crop-ROIHead}
& 50  & 8.8 & 0\% & 0\%  & \multirow{3}{*}{+35.4} & \multirow{3}{*}{--}\\
& 200 & 42.2 & 50\% & 0\%  &  & \\
& 800 & 44.1 & 30\% & 0\%  &  & \\
\bottomrule
\end{tabular}
\end{adjustbox}
\end{table}

\begin{figure}[!ht]\centering
\includegraphics[width=0.90\linewidth]{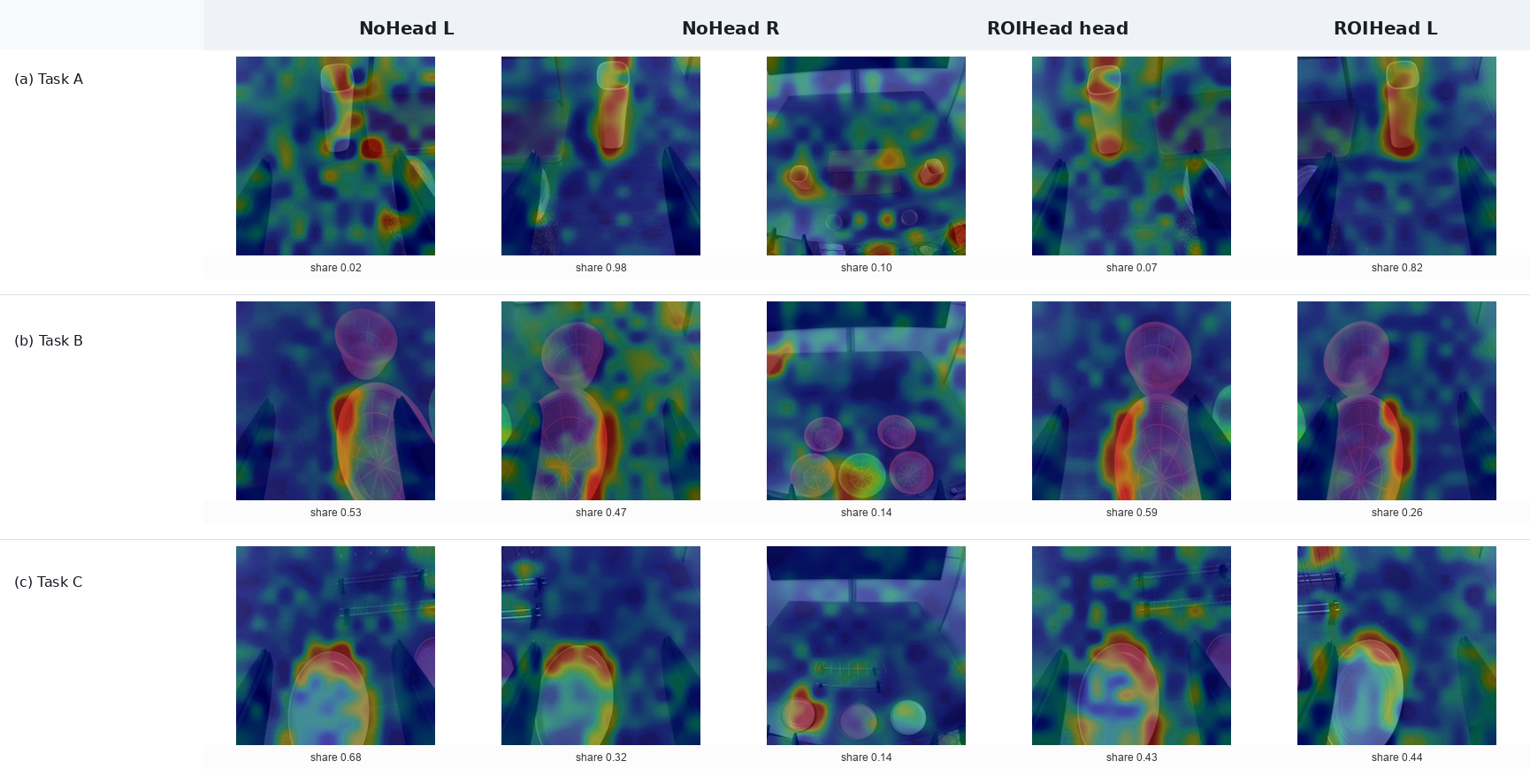}
\caption{RISE-style action-sensitivity comparison between NoHead and ROIHead under the restricted-wrist-FoV setting. NoHead contains only the two wrist views, while ROIHead adds the head view. In these representative single-step examples, wrist views have the larger $q_v$ values, while the head view provides auxiliary spatial cues.}
\label{fig:heatmap-nohead-roi}
\end{figure}

\clearpage
\section{Supplementary Toy-Transfer Diagnostic Details}
\label{app:toy-diagnostic}
\setcounter{table}{0}\setcounter{figure}{0}
This appendix provides the action-sensitivity and representative behavior examples for the toy-transfer diagnostic in Section~\ref{sec:supp-diagnostic}.

\begin{figure}[!ht]\centering
\includegraphics[width=\linewidth]{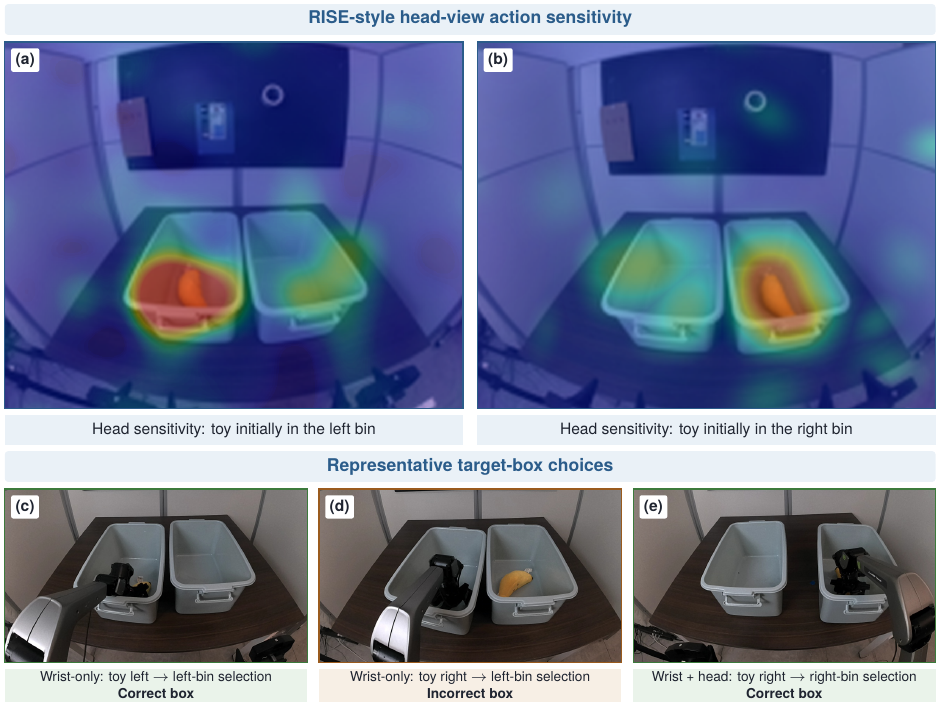}
\caption{RISE-style head-view action sensitivity and representative target-box choices in the toy-transfer diagnostic. (a,b) For the two initial toy locations, the dominant high-response region overlaps the occupied bin and shifts with the visible toy location. The overlays visualize $S_v(c)$ and are interpreted spatially; color intensity is not compared quantitatively across panels. (c,d) In the shown wrist-only cases, the policy selects the left bin for both toy locations, yielding a correct box choice when the toy is on the left and an incorrect choice when it is on the right. (e) In the shown case with an added head view, the policy selects the right bin when the toy is on the right. The behavior frames are recorded from the head-camera viewpoint for visualization. These unscored examples do not estimate success rates, prove causal reliance, or establish the cause of the wrist-only left-bin preference.}
\label{fig:toy-transfer-sensitivity-behavior}
\end{figure}

\clearpage
\section{Training Hyperparameters}
\label{app:training}
\setcounter{table}{0}\setcounter{figure}{0}
All configurations share the same optimization recipe: AdamW ($\beta=(0.9,0.95)$, $\epsilon=10^{-8}$, weight decay $10^{-10}$), global-norm gradient clipping at 1.0, and a warmup-cosine learning-rate (LR) schedule with a 1k-step warmup, peak LR $2.5\times10^{-5}$, and final LR $2.5\times10^{-6}$ after 30k decay steps.

\section{Action-Sensitivity Formulation}
\label{app:rise}
\setcounter{table}{0}\setcounter{figure}{0}
\renewcommand{\theequation}{\thesection\arabic{equation}}\setcounter{equation}{0}
This appendix gives the formulation of the RISE-style action-sensitivity diagnostic summarized in Section~\ref{sec:action-sensitivity}.

For a selected rollout step, let $\mathbf{o}$ be the policy observation and let $\mathbf{a}_0=f_\theta(\mathbf{o})$ be the unmasked action chunk.
For camera view $v$, we sample $K$ binary masks $\{\mathbf{m}_k\}_{k=1}^{K}$ on a $14\times14$ grid, with each cell independently kept with probability $p=0.5$.
The standard diagnostic uses $K=512$ masks; selected paper-rendered examples may use a larger $K$ with the same estimator.
Each mask is upsampled to the image resolution by nearest-neighbor expansion.
Masked pixels are replaced by the mean color of the same image:
\begin{equation}
\tilde{\mathbf{x}}_{v,k}
=
\mathbf{m}_k\odot \mathbf{x}_v
+(1-\mathbf{m}_k)\odot \boldsymbol{\mu}_v .
\end{equation}
The policy is then re-evaluated with only view $v$ perturbed, producing $\mathbf{a}_{v,k}$.
Let $\boldsymbol{\sigma}_{\epsilon}=\max(\boldsymbol{\sigma},\epsilon\mathbf{1})$ elementwise, with $\epsilon=0.01$.
We define the normalized action deviation
\begin{equation}
d_{v,k}
=
\left\|
\frac{\mathbf{a}_{v,k}-\mathbf{a}_0}{\boldsymbol{\sigma}_{\epsilon}}
\right\|_2,
\label{eq:action-sensitivity-delta}
\end{equation}
where $\boldsymbol{\sigma}$ is the per-dimension action standard deviation estimated from valid records.
The action-sensitivity score for grid cell $c$ is the average deviation when that cell is occluded:
\begin{equation}
S_v(c)
=
\frac{\sum_{k=1}^{K} (1-m_k(c))\,d_{v,k}}
{\max\left(\sum_{k=1}^{K} (1-m_k(c)),\,\eta\right)}.
\label{eq:rise-action-sensitivity}
\end{equation}
For view-level summaries, we also compute $D_v$ by masking all cells in view $v$ and measuring the same normalized action deviation.
The view share shown in the diagnostic overlays is
\begin{equation}
q_v =
\begin{cases}
\dfrac{D_v}{\sum_{u\in\mathcal{V}}D_u}, &
\sum_{u\in\mathcal{V}}D_u > 0,\\
0, & \text{otherwise}.
\end{cases}
\label{eq:view-share}
\end{equation}
Here $\eta$ is a small numerical stabilizer for empty cell-level denominators, and $\mathcal{V}$ denotes the valid camera views at the selected step.

\end{document}